\documentclass{article}

\usepackage{graphicx}
\usepackage{booktabs}

\usepackage{hyperref}

\usepackage{amsmath}
\usepackage{amsfonts}
\usepackage{amsthm}
\usepackage{bbm}
\usepackage{enumitem}
\usepackage{subcaption}
\usepackage{placeins}
\usepackage{array}
\usepackage{nicefrac}
\usepackage{natbib}
\newcolumntype{P}[1]{>{\centering\arraybackslash}p{#1}}

\DeclareMathOperator*{\WpEnt}{Ent}
\newcommand*\condPost{\relax\ifmmode(\dagger)\else$(\dagger)$\fi}
\newcommand*\condA{\relax\ifmmode(\star)\else$(\star)$\fi}
\newcommand*\condPrior{\relax\ifmmode(\sharp)\else$(\sharp)$\fi}
\newcommand*\condPriorP{\relax\ifmmode(\sharp')\else$(\sharp')$\fi}

\renewcommand{\epsilon}{\varepsilon}

\newcommand*\bmu{\boldsymbol{\mu}}
\newcommand*\bnu{\boldsymbol{\nu}}
\newcommand*\bsigma{\boldsymbol{\sigma}}
\newcommand*\bSigma{\boldsymbol{\Sigma}}
\newcommand*\btheta{\boldsymbol{\theta}}

\theoremstyle{plain}
\newtheorem{theorem}{Theorem}
\newtheorem{proposition}{Proposition}
\newtheorem{corollary}{Corollary}
\newtheorem{lemma}{Lemma}
\newtheorem{remark}{Remark}
\newtheorem{assumption}{Assumption}

\newtheorem{example}{Example}

\newenvironment{proof2}{\textit{Intuition.}}{\hfill$\square$}
\newtheorem{postulate}{Postulate}

\makeatletter
\newcommand*{\corenum}[1]{%
	\expandafter\@corenum\csname c@#1\endcsname%
}

\newcommand*{\@corenum}[1]{%
	$\ifcase#1\or{\text{iii}}\or{\text{iv}}%
	\else\@ctrerr\fi$%
}
\AddEnumerateCounter{\corenum}{\@corenum}{iv}
\makeatother

\makeatletter
\newcommand*{\condcp}[1]{%
	\expandafter\@condcp\csname c@#1\endcsname%
}

\newcommand*{\@condcp}[1]{%
	$\ifcase#1\or{\text{ii'}}%
	\else\@ctrerr\fi$%
}
\AddEnumerateCounter{\condcp}{\@condcp}{ii'}
\makeatother

\makeatletter
\def\@footnotecolor{red}
\define@key{Hyp}{footnotecolor}{%
	\HyColor@HyperrefColor{#1}\@footnotecolor%
}
\def\@footnotemark{%
	\leavevmode
	\ifhmode\edef\@x@sf{\the\spacefactor}\nobreak\fi
	\stepcounter{Hfootnote}%
	\global\let\Hy@saved@currentHref\@currentHref
	\hyper@makecurrent{Hfootnote}%
	\global\let\Hy@footnote@currentHref\@currentHref
	\global\let\@currentHref\Hy@saved@currentHref
	\hyper@linkstart{footnote}{\Hy@footnote@currentHref}%
	\@makefnmark
	\hyper@linkend
	\ifhmode\spacefactor\@x@sf\fi
	\relax
}%
\makeatother
\hypersetup{colorlinks=true,citecolor=blue, urlcolor=blue, linkcolor=blue, filecolor=black,footnotecolor=red}

\title{An Equivalence between Bayesian Priors and Penalties in Variational Inference}

\author{Pierre Wolinski%
	\thanks{Laboratoire de Math\'ematiques d'Orsay,
		Universit\'e Paris-Saclay, France,
		\texttt{pierre.wolinski@normalesup.org}} 
	\and Guillaume Charpiat\thanks{Tau Team, LRI, Inria Saclay, Saclay, France} 
	\and Yann Ollivier\thanks{Meta AI Research, Paris, France}}

\date{\vspace*{-.4cm}}

\begin{document} \sloppy
	
\maketitle

\begin{abstract}
	In machine learning, it is common to optimize the parameters
	of a probabilistic model, modulated by an ad hoc regularization term that
	penalizes some values of the parameters. Regularization
	terms appear naturally in Variational Inference, a tractable way
	to approximate
	Bayesian posteriors: the loss to optimize contains a
	Kullback--Leibler divergence term between the
	approximate posterior and a Bayesian prior.
	We fully characterize the regularizers that can arise 
	according to this procedure, and
	provide a systematic way to compute the prior
	corresponding to a given penalty. 
	Such a characterization can be used to discover 
	constraints over the penalty function,
	so that the overall procedure remains Bayesian.
\end{abstract}

\section{Introduction}

Adding a penalty term to a loss, in order to make the trained model fit some user-defined property, is very 
common in machine learning. For instance, penalties are used to improve generalization, 
encourage sparsity in a model, or prune neurons 
and reduce the rank of the tensors of weights in the case of neural networks.
Therefore, usual penalties are user-defined and justified mostly empirically, 
and integrated to the loss as follows:
\begin{align*}
\mathcal{L}(\btheta) &= \ell(\btheta) + r(\btheta) ,
\end{align*}
with $\btheta$ the vector of parameters in the model, $\ell(\btheta)$ the error term and 
$r(\btheta)$ the penalty.

From a Bayesian point of view, optimizing such a loss $\mathcal{L}$ is equivalent to finding the Maximum A Posteriori 
(MAP) of the parameters $\btheta$ given the training data and a prior 
$\alpha(\btheta) \propto \exp (- r(\btheta))$. 
Indeed, assuming that the loss $\ell$ is a log-likelihood loss, namely,
$\ell(\btheta)=-\ln p_{\btheta}(\mathcal{D})$ with dataset
$\mathcal{D}$, then
minimizing $\mathcal{L}$ is equivalent to minimizing $
\mathcal{L}_{\mathrm{MAP}}(\btheta) = -\ln p_{\btheta}(\mathcal{D}) - 
\ln(\alpha(\btheta))$.
Thus, within the MAP framework, we can interpret the penalty term $r$ as
the influence of a prior $\alpha$ \citep{mackay1992practical}.

However, the MAP approximates the Bayesian posterior very roughly, by taking its maximum. 
Variational Inference (VI) provides a \emph{variational posterior} distribution rather than a
single value, hopefully representing the Bayesian posterior much better.
VI looks for the best posterior approximation within a \emph{variational family}
$(\beta_{\mathbf{u}}(\btheta))_{\mathbf{u}}$ of approximate posteriors over
$\btheta$, parameterized by a vector $\mathbf{u}$. 
So, the \emph{variational parameters} $\mathbf{u}$ are trained instead of $\btheta$.
Typically, one can choose, for $\btheta \in \mathbb{R}^N$,
$\bmu = (\mu_1, \cdots, \mu_N)$ and $\bsigma = (\sigma_1, \cdots, \sigma_N)$:
\begin{align*}
	\beta_{\mathbf{u}} = \beta_{\bmu, \bsigma} = 
	\mathcal{N}(\mu_1, \sigma_1^2) \otimes \cdots \otimes \mathcal{N}(\mu_N, \sigma_N^2) .
\end{align*}

In VI, the loss to be minimized over $\mathbf{u}$ is:
\begin{align}
\label{eq:introVI}
\mathcal{L}_{\mathrm{VI}}(\beta_{\mathbf{u}}) & \;=\; -\mathbb{E}_{\btheta \sim 
		\beta_{\mathbf{u}}} \ln
		p_{\btheta}(\mathcal{D}) \,+\, 
\mathrm{KL}(\beta_{\mathbf{u}} \| \alpha) \nonumber \\
&\;=\; \underbrace{\mathbb{E}_{\btheta \sim 
		\beta_{\mathbf{u}}} \ell(\btheta)
		}_{\text{data fit term}} \,+\, 
\underbrace{ \mathrm{KL}(\beta_{\mathbf{u}} \| \alpha)}_{\text{penalty term}} ,
\end{align}
which is also an upper bound on
the Bayesian negative log-likelihood of the data \citep{jordan1999introduction}.
Here $p_{\btheta}(\mathcal{D})$ is the likelihood of the full dataset
$\mathcal{D}$ given $\btheta$, $\ell(\btheta)=-\ln
p_{\btheta}(\mathcal{D})$ is the log-likelihood loss,
and $\alpha$ is the Bayesian prior. The variational posteriors
$\beta_{\mathbf{u}}$ that minimize this loss will be concentrated around
values of $\btheta$ that assign high probability to the data, while
not diverging too much from the prior $\alpha$.
Thus, the $\mathrm{KL}$-divergence term can be seen as a penalty $r(\mathbf{u})$.

\paragraph{Are we still Bayesian when using a custom penalty?}
We study whether the prior-penalty correspondence above for the MAP extends 
to the richer variational inference setup. We assume that we optimize a VI loss of the type:
\begin{align*}
	\mathcal{L}(\beta_{\mathbf{u}}) = \mathbb{E}_{\btheta \sim \beta_{\mathbf{u}}} 
	\ell(\btheta) + r(\mathbf{u}) ,
\end{align*}
and consider the following question:
given a variational family $(\beta_{\mathbf{u}})_{\mathbf{u}}$
and a penalty $r(\mathbf{u})$, can $r(\mathbf{u})$ be interpreted as 
a $\mathrm{KL}$-divergence $\mathrm{KL}(\beta_{\mathbf{u}} \| \alpha)$
for some prior distribution $\alpha$?
We answer this question in Section \ref{sec:penalty_KL},
with Theorem \ref{thm:main} and Corollary \ref{cor:main},
and we provide a formula for the corresponding prior $\alpha$,
if it exists.
With these theoretical results, it is possible
to check whether a setup 
with a custom penalty $r$ is actually Bayesian.
Moreover, if it is the case, we provide a formula 
to compute the corresponding prior $\alpha$ explicitly.

\paragraph{How to tune the shape of a penalty to make it fit the Bayesian setup?}
Let us consider a variational family with a penalty containing 
several undetermined hyperparameters (possibly functions).
Results of Section \ref{sec:penalty_KL} can be used
to find constraints over these hyperparameters
to make the penalty equal to a $\mathrm{KL}$-divergence with a
prior $\alpha$. That is, we provide a hyperparameter
tuning method based on Bayesian principles,
and we use it in several examples in Section \ref{sec:examples}.


Our main contributions are five-fold:
\begin{enumerate}
	\item we show how to interpret a custom penalty as the influence of a Bayesian prior,
	if possible, and we propose a formula to compute the prior (Theorem \ref{thm:main});
	\item under some conditions over the variational family and
	the search space of the prior, we provide a set of necessary and sufficient
	conditions that the penalty must fulfill to ensure existence and uniqueness
	of a corresponding prior (Corollary \ref{cor:main});
	\item we apply these results to a series of examples, 
	showing that, among other things: 
	if the penalty is $\mathcal{L}^2$ and corresponds 
	to a prior, then this prior is Gaussian (Corollary \ref{cor:appl:l2_penalty}); 
	if the penalty is $\mathcal{L}^1$ and the candidate variational posteriors
	have a smooth density, then no corresponding prior exists (Proposition \ref{cor:penalty_L1});
	\item additionally, we provide a result about sums
	of penalties, when each one involves the mean of a different 
	parameter: if this sum corresponds to a Bayesian prior, then
	this prior can be decomposed into a product of independent priors, each
	one over a different parameter (Corollary \ref{cor:indep_pen});
	\item we show that these theoretical results can
	be used to tune the strength of the penalty when training neural networks
	in a VI setup.
\end{enumerate}

\paragraph{Outline.} 
Section \ref{sec:related}, provides a short review of the related works.
In Section \ref{sec:varinf}, we recall the general variational inference 
setup.
Then, we prove the main results in Section \ref{sec:penalty_KL}:
we provide a reminder on the theory of distributions in Section \ref{sec:distr},
which is necessary to prove the main theorem in Section \ref{sec:main_result}, and then,
we provide some corollaries concerned with sums of penalties in 
Section \ref{sec:sum_penalties}.
Afterwards, in Section \ref{sec:examples}, we show how our theoretical results can be used
in a series of examples.
We also provide in Section \ref{sec:application} an application of
the preceding results to the construction of a heuristic for 
the strength of the penalty.
Finally, we discuss the hypotheses and possible extensions
of the main theoretical results in Section \ref{sec:discussion}.

\section{Related Work} \label{sec:related}

Interpretations of existing empirical deep learning methods
in a Bayesian variational inference framework include, for instance, 
\emph{variational drop-out} \citep{kingma2015variational},
a version of drop-out which fits the Bayesian framework. 
Further developments of variational drop-out have been made 
by \cite{molchanov2017variational} for weight pruning and by \cite{louizos2017bayesian} for neuron pruning. 

Closer to our present work, the links between a penalized loss and the Bayesian point of view have previously 
been mentioned by \cite{olshausen1997sparse} and used to favor sparsity, but only 
within the MAP approximation.
\cite{mackay1992practical} noted the equivalence  of the penalized loss
$\mathcal{L}(\btheta) = \ell(\btheta) + r(\btheta)$  and the MAP loss
$\mathcal{L}_{\mathrm{MAP}}(\btheta) = \ell(\btheta) -
\ln(\alpha(\btheta))$ 
when $\ell(\btheta)$ is the negative log-likelihood of the training dataset given the parameters $\btheta$, 
and with a prior $\alpha(\btheta) \propto \exp(- r(\btheta))$.
That is, finding the vector $\hat{\btheta}$ minimizing a loss $\mathcal{L}$ 
can be equivalent to finding the MAP estimator $\hat{\btheta}_{\mathrm{MAP}}$ 
by minimizing the loss $\mathcal{L}_{\mathrm{MAP}}$ with a well-chosen prior 
distribution $\alpha$.
In the same vein, \cite{gribonval2011should} has studied the Bayesian interpretation of the
penalized least squares problem, and found out that a single penalty term
can be interpreted as the influence of different priors, 
depending on the choice of the estimator.%
\footnote{\cite{gribonval2011should} focused on
the MAP and the Minimum Mean-Square Error (MMSE) estimators.}

However, the MAP framework is not completely satisfying from a Bayesian point of view: 
instead of returning a distribution over the parameters, 
which contains information about their uncertainty, 
it returns the most reasonable  value. 
In order to evaluate this uncertainty, \cite{mackay1992bayesian} proposed a second-order 
approximation of the Bayesian posterior. 
In the process, \cite{mackay1995probable} also proposed a complete Bayesian 
framework and interpretation of neural networks. Still, this approximation of the 
Bayesian posterior is quite limited.
In the same period, \cite{hinton1993keeping} applied the Minimum Description Length 
(MDL) principle to neural networks. Then, \cite{mackay2003information} made the 
link between the MDL principle and variational inference, and \cite{graves2011practical} applied it to  
neural networks, allowing for variational approximations of the Bayesian
posterior in a tractable way.

In a different perspective, variational inference is a key ingredient of
the Variational Auto-Encoders (VAE) \citep{kingma2014auto}
and subsequent variations and improvements, such as generative diffusion models
\citep{ho2020denoising}. 
Given a distribution $\mathrm{P}$ of data points lying in a high-dimensional
space (e.g., space of images), VAE models are 
trained to embed their inputs into a low-dimensional space 
(\emph{encoding of the inputs})
and reconstruct these inputs by using only their embeddings
(\emph{decoding of the embeddings}).
Such models can then be used to:
1) compress samples from $\mathrm{P}$ by encoding them;
2) generate new samples from $\mathrm{P}$ by decoding random noise.
In the VAE setup, the loss is penalized by a $\mathrm{KL}$-divergence
between the distribution of the embeddings and a
Bayesian prior, similarly to Eqn.~\eqref{eq:introVI}.
In this context, replacing the $\mathrm{KL}$ penalty term by 
a custom penalty is still under discussion \citep[Appendix G]{rombach2022high}.
So, identifying conditions that a custom penalty must fulfill
to correspond to the influence of a Bayesian prior remains
a problem to be solved in the VAE training framework.

Moreover, variational inference is a way of approximating the Bayesian posterior which presents several 
advantages. First, as recalled by \cite{knoblauch2019generalized}, the penalty used in variational inference 
is optimal for approximating the true Bayesian posterior. That is, any other penalty performs worse. Second, 
variational inference allows us to compute the variational posterior by gradient descent, which is a major 
advantage when the model to train is a neural network.

Additionally, a series of recent 
papers \citep{alquier2020concentration,cherief2020convergence,zhang2020convergence} provide theoretical results 
about the convergence speed of the variational posterior, according to the size of the training set. 
Therefore, translating user-specified penalties into Bayesian priors is a way to fit a framework in which 
several results are already known.


\section{Variational Inference} \label{sec:varinf}

We propose here a reminder on variational inference,
following \cite{graves2011practical}.

From a Bayesian viewpoint, we describe the vector of parameters $\btheta \in \mathbb{R}^N$ of a model 
as a random variable. 
Given a dataset $\mathcal{D}$, we denote by
$p_{\btheta}(\mathcal{D})$ the likelihood of $\mathcal{D}$ given $\btheta$, that is, 
the probability that the model with parameter $\btheta$ 
assigns to $\mathcal{D}$.
For instance, with a dataset $\mathcal{D} = \{(x_1, y_1),
\cdots , (x_n, y_n) \}$ of $n$ input-output pairs and a model that
assigns probabilities $p_{\btheta}(y_i | x_i)$ for the outputs, then $\ln
p_{\btheta}(\mathcal{D})=\sum_{i=1}^n \ln p_{\btheta}(y_i | x_i)$
is the total $\log$-likelihood of the data given the model.

Given the dataset $\mathcal{D}$, the posterior distribution
over the parameters $\btheta$ is:
\begin{align*}
\pi_{\mathcal{D}}(\btheta) = \frac{p_{\btheta}(\mathcal{D}) \alpha(\btheta)}{\mathbb{P}(\mathcal{D})} 
,\quad \mathbb{P}(\mathcal{D})=\int_{\mathbb{R}^N}
p_{\btheta}(\mathcal{D}) \alpha(\btheta) \, \mathrm{d}{\btheta} ,
\end{align*}
which is analytically intractable for complicated models, 
such as multi-layer nonlinear neural networks.
However, the posterior $\pi_D$ can be approximated by looking for
probability distributions $\beta$ that minimize the loss:
\begin{align}
\mathcal{L}_{\mathrm{VI}}(\beta) = 
- \mathbb{E}_{\btheta \sim \beta} \ln 
p_{\btheta} (\mathcal{D}) + \mathrm{KL}(\beta \| \alpha)  \label{eqn:varinf_base} ,
\end{align}
where 
$\mathrm{KL}(\beta \| \alpha) = \int_{\mathbb{R}^N} \ln( \frac{\beta(\btheta)}{\alpha(\btheta)}) 
\beta(\btheta) \, \mathrm{d} \btheta$ is the
Kullback--Leibler divergence. 
Indeed, one has $\mathcal{L}_{\mathrm{VI}}(\beta) = -\ln
\mathbb{P}(\mathcal{D})+\mathrm{KL}(\beta\|\pi_{\mathcal{D}})$, which is
minimized with $\beta=\pi_{\mathcal{D}}$.
Besides, one can decompose the $\mathrm{KL}$-divergence into 
the \emph{entropy} term and the \emph{cross-entropy} term:
\begin{align*}
	\mathrm{KL}(\beta \| \alpha) = -
	\underbrace{\left[-\int_{\mathbb{R}^N} \ln( \beta(\btheta)) 
		\beta(\btheta) \, \mathrm{d} \btheta
	\right]}_{\text{entropy } \WpEnt(\beta) \text{ of } \beta}
	+ \underbrace{\left[-\int_{\mathbb{R}^N} \ln(\alpha(\btheta)) 
	\beta(\btheta) \, \mathrm{d} \btheta
	\right]}_{\text{cross-entropy between } \beta \text{ and } \alpha} .
\end{align*}

The first term in the loss \eqref{eqn:varinf_base} represents the
error made over the dataset $\mathcal{D}$: it is small if $\beta$ is
concentrated around good parameters $\btheta$.
The second term
can be seen as a user-defined penalty over $\beta$ that keeps it
from diverging too much from the prior $\alpha$.
Moreover, for any distribution $\beta$, the
quantity $\mathcal{L}_{\mathrm{VI}}(\beta)$ is a upper bound on the Bayesian
negative log-likelihood of the data: $\mathcal{L}_{\mathrm{VI}}(\beta)\geq -\ln \int_{\mathbb{R}^N}
\alpha(\btheta)p_{\btheta}(\mathcal{D}) \, \mathrm{d} \btheta$
\citep{jordan1999introduction}.

In variational inference, a parametric family $(\beta_{\mathbf{u}})_{\mathbf{u}}$ of probability
distributions is fixed, and one looks for the best approximation
$\beta_{\mathbf{u}^*}$ of the Bayesian posterior by minimizing
$\mathcal{L}_{\mathrm{VI}}(\beta_{\mathbf{u}})$ according to $\mathbf{u}$.
So, $(\beta_{\mathbf{u}})_{\mathbf{u}}$ is the \emph{variational family},
$\mathbf{u}$ is the \emph{variational parameter}
and $\beta_{\mathbf{u}^*}$ is the \emph{variational posterior}. 
Importantly, for some families
such as Gaussians with fixed variance, the gradient of
$\mathcal{L}_{\mathrm{VI}}(\beta_{\mathbf{u}})$ can be computed if the gradients of $\ln
p_{\btheta}(\mathcal{D})$ can be computed \citep{graves2011practical}, so that
$\mathcal{L}_{\mathrm{VI}}(\beta_{\mathbf{u}})$ can be optimized by stochastic gradient descent. 
Thus, this is well-suited for models such as neural networks.

To summarize, we consider a parametric family $(\beta_{\mathbf{u}})_{\mathbf{u}}$
with parameter $\mathbf{u}$, where each
$\beta_{\mathbf{u}}$ is a probability distribution over $\btheta \in \mathbb{R}^N$.
Then we learn the variational parameters $\mathbf{u}$ instead of $\btheta$. For instance, 
we can choose one of the following families of variational posteriors.

\begin{example} \label{ex:gaussian}
	The family of products of Gaussian distributions over $\btheta = (\theta_1, \cdots, 
	\theta_N)$:
	\begin{align*}
	\beta_{\mathbf{u}} = \beta_{\bmu, \bsigma}
	= \mathcal{N}(\mu_1, \sigma_1^2) \otimes \cdots \otimes \mathcal{N}(\mu_N, \sigma_N^2) .
	\end{align*}
	In this case, the parameters of the model estimated by VI
	are random and independently sampled from different 
	Gaussian distributions $\mathcal{N}(\mu_k, \sigma_k^2)$. Instead of learning them directly, the vector of 
	parameters $\mathbf{u} = (\mu_1, \sigma_1, \cdots , \mu_N,
	\sigma_N)$ is learned in such a way that they minimize
	$\mathcal{L}_{\mathrm{VI}}(\beta_\mathbf{u})$.
\end{example}

\begin{example} \label{ex:dirac}
	The family of products of Dirac distributions over $\btheta = (\theta_1, \cdots, 
	\theta_N)$:
	\begin{align*}
	\beta_{\mathbf{u}} = \beta_{\bmu} = \delta_{\mu_1} \otimes \cdots \otimes 
	\delta_{\mu_N}.
	\end{align*}
	In this case, the parameters are deterministic: $\theta_k$ and $\mu_k$ are identical for all $k$.
\end{example}

\section{Bayesian Interpretation of Penalties} \label{sec:penalty_KL}

In this section, we provide conditions
ensuring that a penalty $r(\mathbf{u})$ over the parameters of a
variational posterior can be 
interpreted as a Kullback--Leibler divergence with respect to a prior
$\alpha$; namely, that:
\begin{align}
\exists \, K\in \mathbb{R},\, \forall \mathbf{u}, \quad r(\mathbf{u}) + K = \mathrm{KL}(\beta_{\mathbf{u}} \| 
\alpha), \label{eqn:main_eqn}
\end{align}
where the constant $K$ is irrelevant for optimization.
In the process, we give a 
formula expressing $\alpha$ as a function of $r(\cdot)$.

More precisely, under some conditions over the variational family (Conditions
\ref{cond:a-:translation}-\ref{cond:b:nonzero}) and
the search space of the prior (space $\mathcal{T}(\mathbb{R}^N)$), 
we provide a set of necessary and sufficient
conditions (Conditions \ref{cond:c-:schwartz}-\ref{cond:d':proper})
that the penalty must fulfill to ensure existence and uniqueness
of a corresponding prior.
The proof is given in Section \ref{sec:main_result} and
is divided into two parts: in Theorem \ref{thm:main},
we state some regularity conditions on the variational family $(\beta_{\mathbf{u}})_{\mathbf{u}}$
and the penalty $r(\mathbf{u})$, so that a candidate prior $\exp(A_{\mathbf{u}})$ would emerge;
in Corollary \ref{cor:main}, we state conditions over $A_{\mathbf{u}}$ so that
$\exp(A_{\mathbf{u}})$ corresponds effectively to the density of a prior.
Then, in Section \ref{sec:sum_penalties}, we 
prove that the prior exhibits a specific structure
when the corresponding penalty is made of a sum of penalties, each one over
a different subset of the parameters.


Since some notions of theory of distributions 
are necessary to our derivation, we start by providing a reminder.

\subsection{Reminder on Theory of Distributions} \label{sec:distr}

In order to explain the main result, we recall some basic concepts of theory of distributions. 
We use three functional spaces: 
the Schwartz class $\mathcal{S}(\mathbb{R}^N)$, the space of tempered distributions 
$\mathcal{S}'(\mathbb{R}^N)$, and the space of distributions with compact support 
$\mathcal{E}'(\mathbb{R}^N)$.

We recall the definition of the space of distributions $\mathcal{D}'(\mathbb{R}^N)$. We denote by 
$C^{\infty}(\mathbb{R}^N)$ the space of infinitely derivable functions mapping $\mathbb{R}^N$ to $\mathbb{R}$, 
and by $C^{\infty}_c(\mathbb{R}^N) \subset C^{\infty}(\mathbb{R}^N)$ the subspace of functions with compact 
support, that is $\varphi \in C^{\infty}_c(\mathbb{R}^N)$ if, and only if:
\begin{gather*}
\varphi \in C^{\infty}(\mathbb{R}^N) \text{ and } \exists K \subset \mathbb{R}^N \text{ compact s.t.: }
\{ x \in \mathbb{R}^N : \varphi(x) \neq 0\} \subseteq K .
\end{gather*}
The set $\{ x \in \mathbb{R}^N : \varphi(x) \neq 0\}$ is also denoted by $\mathrm{supp}(\varphi)$.

\paragraph{Space of distributions $\mathcal{D}'(\mathbb{R}^N)$.} The space of distributions 
$\mathcal{D}'(\mathbb{R}^N)$ is defined as the space of continuous linear forms over 
$C^{\infty}_c(\mathbb{R}^N)$. For any \emph{distribution} $T \in \mathcal{D}'(\mathbb{R}^N)$, we denote by 
$\langle T, \phi \rangle$ the value of $T$ at a given \emph{test function} $\varphi \in 
C^{\infty}_c(\mathbb{R}^N)$.

More formally, $T \in \mathcal{D}'(\mathbb{R}^N)$ if, and only if, for all compact set $K$ of $\mathbb{R}^N$, 
there exists $p \in \mathbb{N}$ and $C > 0$ such that:
\begin{gather*}
\forall \varphi \in C^{\infty}_c(\mathbb{R}^N) \text{ with } \mathrm{supp}(\varphi) \subseteq K, \\
|\langle T, \varphi \rangle| \leq C \sup_{|\alpha| \leq p} \|\partial^{\alpha} \varphi\|_{\infty} ,
\end{gather*}
where $\alpha \in \mathbb{N}^{N}$ is a multi-index representing the order of derivation
according to each of the $N$ variables,
and $|\alpha| = \sum_{i = 1}^N \alpha_i$.

Distributions are easier to visualize in specific cases. For instance, if a function $f : \mathbb{R}^N 
\rightarrow \mathbb{R}$ is integrable on every compact set $K \in \mathbb{R}^N$, then we can define a 
distribution $T_f$ by:
\begin{align*}
\forall \varphi \in C_c^{\infty}(\mathbb{R}^N), \quad \langle T_f, \varphi \rangle = \int f \varphi .
\end{align*}
Therefore, distributions are often called ``generalized functions'', since most functions can be seen as 
distributions. In fact, by abuse of notation, $\langle f, \varphi \rangle$ stands for $\langle T_f, \varphi 
\rangle$.

Another classic example of distribution is the Dirac at zero $\delta$, defined as follows:
\begin{align*}
\forall \varphi \in C_c^{\infty}(\mathbb{R}^N), \quad \langle \delta, \varphi \rangle = \varphi(0) .
\end{align*}

Finally, the derivatives $\partial^{\alpha} T$ of a distribution $T$ are defined as follows:
\begin{align*}
	\forall \varphi \in C^{\infty}_c(\mathbb{R}^N), \quad 
	\langle \partial^{\alpha} T, \varphi \rangle = (-1)^{|\alpha|} \langle T, \partial^{\alpha} \varphi \rangle .
\end{align*}
Notably, when $N = 1$, $\langle T', \varphi \rangle = - \langle T, \varphi' \rangle$.

\paragraph{Schwartz class $\mathcal{S}(\mathbb{R}^N)$.} $\varphi$ belongs to 
the Schwartz class $\mathcal{S}(\mathbb{R}^N)$, if and only if,
$\varphi \in C^{\infty}(\mathbb{R}^N)$ and:
\begin{align*}
\forall p \in \mathbb{N}, \exists C_p > 0 : \sup_{|\alpha| \leq p , |\beta| \leq p} \|x^{\alpha} 
\partial^{\beta} \varphi(x)\|_{\infty} \leq C_p .
\end{align*}
In other words, $\mathcal{S}(\mathbb{R}^N)$ contains smooth and rapidly decreasing functions. For instance, any 
Gaussian density function belongs to $\mathcal{S}(\mathbb{R}^N)$.

We can easily define the Fourier transform on $\mathcal{S}(\mathbb{R}^N)$. Let $\varphi \in 
\mathcal{S}(\mathbb{R}^N)$:
\begin{align*}
(\mathcal{F}\varphi)(\xi) &= \int_{\mathbb{R}^N} \varphi(x) e^{-i \xi x} \, \mathrm{d} x .
\end{align*}
Thus, $\mathcal{F}^{-1} = (2 \pi)^{-N} \bar{\mathcal{F}}$, where $\bar{\mathcal{F}} \varphi = \mathcal{F} 
\check{\varphi}$ and $\check{\varphi}(x) = \varphi(-x)$.

\paragraph{Space of tempered distributions $\mathcal{S}'(\mathbb{R}^N)$.} Let $T \in 
\mathcal{D}'(\mathbb{R}^N)$ be a distribution. $T$ belongs to $\mathcal{S}'(\mathbb{R}^N)$ if, and only if, 
there exists $p \in \mathbb{N}$ and $C > 0$ such that:
\begin{align*}
\forall \varphi \in C^{\infty}_c(\mathbb{R}^N), \;\; |\langle T, \varphi \rangle| \leq \sup_{|\alpha| \leq 
	p, |\beta| \leq p} \|x^{\alpha} \partial^{\beta} \varphi \|_{\infty} .
\end{align*}

The Fourier transform is defined on $\mathcal{S}'(\mathbb{R}^N)$ by duality. For any $T \in 
\mathcal{S}'(\mathbb{R}^N)$, $\mathcal{F} T \in \mathcal{S}'(\mathbb{R}^N)$ and is defined by:
\begin{align*}
\forall \varphi \in \mathcal{S}(\mathbb{R}^N), \quad \langle \mathcal{F} T, \varphi \rangle = \langle T, 
\mathcal{F} 
\varphi \rangle .
\end{align*}

Notably, this definition allows us to compute the Fourier transform of functions that do not lie in 
$\mathcal{L}^2$. This is very useful in the applications of Theorem~\ref{thm:main}, where we need the Fourier 
transform of $f : x \mapsto x^2$, which is $\mathcal{F}f = - 2 \pi \delta'' \in \mathcal{S}'(\mathbb{R})$ 
(where $\delta''$ is the second derivative of the Dirac, defined below).

\paragraph{Space of distributions with compact support $\mathcal{E}'(\mathbb{R}^N)$.} Let $T \in 
\mathcal{D}'(\mathbb{R}^N)$. The support of $T$ is defined by:
\begin{align*}
\mathrm{supp}(T) = \mathbb{R}^N \backslash \{ x \in \mathbb{R}^N : \exists \omega \in \mathcal{N}_x \text{ 
	s.t.\ } T_{| \omega} = 0 \} ,
\end{align*}
where $\mathcal{N}_x$ is the set of neighborhoods of $x$.
Thus, $T$ is said to have a compact support if, and only if, $\mathrm{supp}(T)$ is contained into a compact 
subset of $\mathbb{R}^N$. As fundamental property of $\mathcal{E}'(\mathbb{R}^N)$, one should notice that 
$\mathcal{E}'(\mathbb{R}^N) \subset \mathcal{S}'(\mathbb{R}^N)$. That is, the Fourier transform is defined on 
$\mathcal{E}'(\mathbb{R}^N)$.

For instance, the Dirac at zero $\delta$ and its derivatives $\delta^{(k)}$ have support $\{0\}$, 
which is compact. For any test function $\varphi \in C^{\infty}_c(\mathbb{R})$:
\begin{align*}
\langle \delta, \varphi \rangle &= \varphi(0), 
&\langle \delta^{(k)}, \varphi \rangle &= (-1)^k \varphi^{(k)}(0) .
\end{align*}

\subsection{Main Results} \label{sec:main_result}

\paragraph{Notation and statement.} In order to approximate the posterior distribution of a vector
$\btheta \in \mathbb{R}^N$, we denote by $(\beta_{\bmu, \bnu})_{\bmu, \bnu}$ the family of 
variational posteriors over $\btheta$, parameterized by their mean $\bmu$ and a vector of additional parameters 
$\bnu$. The basic example is a multivariate Gaussian distribution $\beta_{\bmu, \bnu}$ 
parameterized by its mean $\bmu \in \mathbb{R}^N$ and its covariance matrix $\bnu = \mathbf{\Sigma} \in 
\mathbb{R}^{N \times N}$.

We say that the family $(\beta_{\bmu, \bnu})_{\bmu, \bnu}$ of
variational posteriors is \emph{translation-invariant} if
$\beta_{\bmu, \bnu}(\btheta) = \beta_{0, \bnu}(\btheta - \bmu)$ for
all $\bmu, \bnu, \btheta$.

We denote by $\mathrm{KL}(f \| g)$ the $\mathrm{KL}$-divergence extended to functions $f$ and $g$,
which are not necessarily probability densities:
\begin{align*}
	\mathrm{KL}(f \| g) = \int \ln\left( \frac{f(\btheta)}{g(\btheta)} \right) f(\btheta) \, \mathrm{d}\btheta
\end{align*}

We denote by $r_{\bnu}(\bmu)$ some penalty, applied to the distribution 
$\beta_{\bmu, \bnu}$.

In the following, we always restrict ourselves to priors
$\alpha \in \mathcal{T}(\mathbb{R}^N) = \{\alpha\; \text{ s.t. } \ln(\alpha) \in \mathcal{S}'(\mathbb{R}^N) \}$, i.e.\ 
log-tempered probability distributions, hence the
condition $\ln(\alpha)\in \mathcal{S}'(\mathbb{R}^N)$ in the results below. 

Also, we make the following assumptions on the variational family:

\begin{assumption} \label{assum:1}
	Let $(\beta_{\bmu, \bnu})_{\bmu, \bnu}$ be a family of variational posteriors such that:
	\begin{enumerate}[label = (\alph*)]
		\item $(\beta_{\bmu, \bnu})_{\bmu, \bnu}$ is translation-invariant; \label{cond:a-:translation}
		\item for all $\bnu$, the density $\beta_{0, \bnu}$ has finite entropy 
		and belongs to $\mathcal{S}(\mathbb{R}^N)$; \label{cond:a:schwartz}
		\item for all $\bnu$, $\mathcal{F} \beta_{0, \bnu}$ is nonzero everywhere. \label{cond:b:nonzero}
	\end{enumerate}
\end{assumption}

\begin{theorem} \label{thm:main}
	We assume that the variational family $(\beta_{\bmu, \bnu})_{\bmu, \bnu}$
	fulfills Assumption~\ref{assum:1}.
	Let $r_{\bnu}(\bmu)$ be a penalty over $\beta_{\bmu, \bnu}$. We 
	assume that, for all $\bnu$:
	\begin{enumerate}[label = (\roman*)]
		\item $r_{\bnu} \in \mathcal{S}'(\mathbb{R}^N)$; \label{cond:c-:schwartz}
		\item the quotient $\nicefrac{\mathcal{F} r_{\bnu}}{\mathcal{F} \check{\beta}_{0, \bnu}}$
		belongs to $\mathcal{S}'(\mathbb{R}^N)$, where 
		$\check{\beta}_{0, \bnu}(\btheta) := \beta_{0, \bnu}(-\btheta)$. \label{cond:c:quotient}
	\end{enumerate}
	
	Then, there exists a unique function $A_{\bnu} \in \mathcal{S}'(\mathbb{R}^N)$ 
	such that:
	\begin{align}
		\qquad & r_{\bnu}(\bmu) = \mathrm{KL}(\beta_{\bmu, \bnu} \| \exp(A_{\bnu})), \label{eqn:thm:1} \\
		\text{moreover: } \qquad & A_{\bnu} = - \mathrm{Ent}(\beta_{0, \bnu})\, \mathbbm{1} \, - \, 
		\mathcal{F}^{-1}\left[ 	
		\frac{\mathcal{F} r_{\bnu}}{\mathcal{F} \check{\beta}_{0, \bnu}} \right] , \qquad \label{eqn:thm:2}
	\end{align}
	where $\check{\beta}_{0, \bnu}(\btheta) := \beta_{0, \bnu}(-\btheta)$ and $\mathbbm{1}$ is the 
	constant function equal to $1$.
\end{theorem}

\begin{proof}
	The proof is based on the resolution of a classical integral equation 
	\citep[Section 10.3-1]{polyanin1998handbook} adapted to the wider framework of the theory of distributions. 
	This extension is necessary, since the Fourier transform of the widely-used $\mathcal{L}^2$ penalty $r(x) = x^2$
	is $-2\pi \delta''$, which is a distribution that cannot be expressed as a function.
	
	Let $A_{\bnu}$ be a function in $\mathcal{S}'(\mathbb{R}^N)$. We have:
	\begin{align*}
	\mathrm{KL}(\beta_{\bmu, \bnu} \| \exp(A_{\bnu})) 
	&= -\WpEnt(\beta_{0, \bnu}) - \int_{\mathbb{R}^N} A_{\bnu}(\btheta) \beta_{\bmu, \bnu}(\btheta) 
	\, \mathrm{d}\btheta \\
	&= -\WpEnt(\beta_{0, \bnu}) - \int_{\mathbb{R}^N} A_{\bnu}(\btheta) \check{\beta}_{0, \bnu}(\bmu - \btheta) 
	\, \mathrm{d}\btheta \\
	&= -\WpEnt(\beta_{0, \bnu}) - (A_{\bnu} * \check{\beta}_{0, \bnu})(\bmu) ,
	\end{align*}
	which exists since the convolution between $\check{\beta}_{0, \bnu} \in \mathcal{S}(\mathbb{R}^N)$
	and $A_{\bnu} \in \mathcal{S}'(\mathbb{R}^N)$ is well-defined 
	\citep[Chap.\ I, Theorem 3.13]{stein2016introduction}.
	
	Since $\mathcal{F}$ is an isomorphism on $\mathcal{S}'(\mathbb{R}^N)$ 
	\cite[Chap.\ I.3]{stein2016introduction}, we have:
	\begin{align*}
		r_{\bnu}(\cdot) = \mathrm{KL}(\beta_{\cdot, \bnu} \| \exp(A_{\bnu})) \quad \Leftrightarrow
		\quad \mathcal{F}r_{\bnu} = \mathcal{F}\left[\mathrm{KL}(\beta_{\cdot, \bnu} \| \exp(A_{\bnu}))\right] .
	\end{align*}
	
	Besides, we have:
	\begin{align*}
		\mathcal{F}\left[\mathrm{KL}(\beta_{\cdot, \bnu} \| \exp(A_{\bnu}))\right] 
		&= -(2 \pi)^N \WpEnt(\beta_{0, \bnu}) \delta 
		- (\mathcal{F} A_{\bnu}) \cdot (\mathcal{F} \check{\beta}_{0, \bnu}),
	\end{align*}
	by using $\mathcal{F}\mathbbm{1} = (2 \pi)^N \delta$ 
	and $\mathcal{F}(A_{\bnu} * \check{\beta}_{0, \bnu}) 
	= (\mathcal{F} A_{\bnu}) \cdot (\mathcal{F} \check{\beta}_{0, \bnu})$,
	which holds since $\check{\beta}_{0, \bnu} \in \mathcal{S}(\mathbb{R}^N)$ 
	and $A_{\bnu} \in \mathcal{S}'(\mathbb{R}^N)$.
	See the proof of \cite[Chap.\ I, Theorem 3.18]{stein2016introduction} for more details.
	So:
	\begin{align}
		r_{\bnu}(\cdot) = \mathrm{KL}(\beta_{\cdot, \bnu} \| \exp(A_{\bnu})) \quad ~~\; & \Leftrightarrow &
		\mathcal{F}r_{\bnu} &= -(2 \pi)^N \WpEnt(\beta_{0, \bnu}) \delta 
		- (\mathcal{F} A_{\bnu}) \cdot (\mathcal{F} \check{\beta}_{0, \bnu}) \nonumber \\
		& \Leftrightarrow & \mathcal{F} A_{\bnu} &= 
		-(2 \pi)^N \WpEnt(\beta_{0, \bnu}) \frac{\delta}{\mathcal{F} \check{\beta}_{0, \bnu}}
		- \frac{\mathcal{F}r_{\bnu}}{\mathcal{F} \check{\beta}_{0, \bnu}} \label{eqn:proof:div0} \\
		& \Leftrightarrow & A_{\bnu} &= 
		- \WpEnt(\beta_{0, \bnu}) \mathbbm{1}
		- \mathcal{F}^{-1}\left[\frac{\mathcal{F}r_{\bnu}}{\mathcal{F} \check{\beta}_{0, \bnu}}\right].
		\label{eqn:proof:final}
	\end{align}
	The equivalence with \eqref{eqn:proof:div0} holds since $\mathcal{F} \check{\beta}_{0, \bnu}$
	is assumed to be nonzero everywhere (Condition \ref{cond:b:nonzero}). 
	We obtain \eqref{eqn:proof:final} by using 
	$\frac{\delta}{\mathcal{F} \check{\beta}_{0, \bnu}} = \frac{\delta}{(\mathcal{F} \check{\beta}_{0, \bnu})(0)} = 
	\delta$, since $\check{\beta}_{0, \bnu}$ is a probability distribution,
	and $\mathcal{F}^{-1}[(2 \pi)^N \delta] = \mathbbm{1}$.
	To ensure the equivalence with \eqref{eqn:proof:div0}, 
	we need to check that $\frac{\mathcal{F}r_{\bnu}}{\mathcal{F} \check{\beta}_{0, \bnu}} 
	\in \mathcal{S}'(\mathbb{R}^N)$, which is true by using Condition \ref{cond:c:quotient}, 
	and we use the fact that $\mathcal{F}^{-1}$ is an isomorphism on $\mathcal{S}'(\mathbb{R}^N)$.
\end{proof}

\paragraph{Remarks about Theorem \ref{thm:main}.}
Theorem \ref{thm:main} holds under some technical assumptions.
Even if some of them fail, the formula is still useful to compute a candidate prior $\alpha$
from a penalty $r(\cdot)$. One would have to apply
Equation \eqref{eqn:thm:2} on a penalty $r$ to compute $A_{\bnu}$,
then check that Conditions \ref{cond:d:indep} and \ref{cond:d':proper} hold, 
define the distribution $\alpha \propto \exp(A)$ as in
Eqn.~\eqref{eqn:cor:2}, and finally compute
$\mathrm{KL}(\beta_{\bmu, \bnu} \| \alpha)$ analytically and compare it to $r$.
However, this process does not guarantee the uniqueness of the solution $\alpha$.

\begin{remark} \label{rem:compact}
	In Theorem \ref{thm:main}, Condition \ref{cond:c:quotient} can be replaced by Condition \ref{eqn:cond3p}: 
	\begin{enumerate}[label=(\condcp*)]
		\item $\mathcal{F}r$ belongs to $\mathcal{E}'(\mathbb{R}^N)$ \label{eqn:cond3p},
	\end{enumerate}
	where $\mathcal{E}'(\mathbb{R}^N)$ is the space of distributions with compact support. 
	Indeed:
	\begin{align*}
	\text{Conditions \ref{eqn:cond3p} and \ref{cond:b:nonzero} } \Rightarrow \text{ Condition 
		\ref{cond:c:quotient}.}
	\end{align*}
\end{remark}

It is more convenient to use Condition \ref{eqn:cond3p} when
the penalty $r$ is known to have a Fourier transform with compact support.
For instance, for any integer $p > 0$, the Fourier transform of the penalty $r_{2 p}(x) = x^{2 p}$ is:
\begin{align*}
\mathcal{F}r_{2 p} = (-1)^p \, 2 \pi \, \delta^{(2 p)} \in \mathcal{E}'(\mathbb{R}),
\end{align*}
where $\delta^{(2 p)}$ is the $2 p$-derivative of the Dirac distribution,
which has a compact support. 
Moreover, by linearity of $\mathcal{F}$, any \emph{finite}
linear combination of $(r_{2 p})_{p}$ has also a 
Fourier transform that is in $\mathcal{E}'(\mathbb{R})$.

\paragraph{Consequences for the prior.}
At this point, the situation is:
\begin{align*}
\text{problem to solve: } \qquad & r_{\bnu}(\bmu) + K
= \mathrm{KL}(\beta_{\bmu, \bnu} \| \alpha), \\
\text{problem solved: } \qquad & r_{\bnu}(\bmu)
= \mathrm{KL}(\beta_{\bmu, \bnu} \| \exp(A_{\bnu})) ,
\end{align*}
where $A_{\bnu}$ is defined in Eqn.~\eqref{eqn:thm:2}, $K \in \mathbb{R}$
and $\alpha$ is a probability distribution to find.
Informally, we expect $\alpha$ to be proportional to $\exp(A_{\bnu})$.
So, two conditions have to be taken into account: 
$A_{\bnu}$ should not depend on $\bnu$ and $\exp(A_{\bnu})$
should integrate to $1$.
In order to obtain such an $\alpha$, the two
following results, although immediate, will be helpful.

\begin{lemma} \label{lem:linearity}
	Let $r^{(1)}_{\bnu}(\bmu)$ and $r^{(2)}_{\bnu}(\bmu)$ be two penalties
	fulfilling the conditions of Theorem \ref{thm:main}. 
	We can build the solutions of Eqn.~\eqref{eqn:thm:1} for $r^{(1)}_{\bnu}$ and $r^{(2)}_{\bnu}$
	by introducing $\bar{A}^{(1)}_{\bnu}$ and $\bar{A}^{(2)}_{\bnu}$:
	\begin{align*}
		r^{(1)}_{\bnu}(\bmu) &= \mathrm{KL}(\beta_{\bmu, \bnu} \| 
		\exp[-\WpEnt(\beta_{0, \bnu}) - \bar{A}^{(1)}_{\bnu}]),
		& \bar{A}^{(1)}_{\bnu} &:= \mathcal{F}^{-1}\left[ 	
		\frac{\mathcal{F} r^{(1)}_{\bnu}}{\mathcal{F} \check{\beta}_{0, \bnu}} \right], \\
		r^{(2)}_{\bnu}(\bmu) &= \mathrm{KL}(\beta_{\bmu, \bnu} \| 
		\exp[-\WpEnt(\beta_{0, \bnu}) - \bar{A}^{(2)}_{\bnu}]), 
		& \bar{A}^{(2)}_{\bnu} &:= \mathcal{F}^{-1}\left[ 	
		\frac{\mathcal{F} r^{(2)}_{\bnu}}{\mathcal{F} \check{\beta}_{0, \bnu}} \right],
	\end{align*}
	Thus, for any $K \in \mathbb{R}$, we have:
	\begin{align*}
		r^{(1)}_{\bnu}(\bmu) + K r^{(2)}_{\bnu}(\bmu)
		&= \mathrm{KL}(\beta_{\bmu, \bnu} \| \exp[-\WpEnt(\beta_{0, \bnu})
		- \bar{A}^{(1)}_{\bnu} - K \bar{A}^{(2)}_{\bnu}]) .
	\end{align*}
\end{lemma}

\begin{lemma} \label{lem:normalization}
	Specifically, Lemma \ref{lem:linearity} allows us to show that:
	\begin{enumerate}[label=(\arabic*)]
		\item for $K = 1$ and $r^{(2)}_{\bnu}(\bmu) = f(\bnu)$, we have: \label{lem:normalization:1}
			\begin{align*}
			r^{(1)}_{\bnu}(\bmu) + f(\bnu) &= \mathrm{KL}(\beta_{\bmu, \bnu} \| 
			\exp[- \mathrm{Ent}(\beta_{0, \bnu}) - \bar{A}^{(1)}_{\bnu} - f(\bnu)]) \\
			A_{\bnu}(\btheta) &:= - \mathrm{Ent}(\beta_{0, \bnu}) - \bar{A}^{(1)}_{\bnu}(\btheta) - f(\bnu) .
			\end{align*}
		\item for $r^{(2)}_{\bnu}(\bmu) = 1$, we have: \label{lem:normalization:2}
			\begin{align*}
			r^{(1)}_{\bnu}(\bmu) + K &= \mathrm{KL}(\beta_{\bmu, \bnu} \| 
			\exp[- \mathrm{Ent}(\beta_{0, \bnu}) - \bar{A}^{(1)}_{\bnu} - K]) \\
			A_{\bnu}(\btheta) &:= - \mathrm{Ent}(\beta_{0, \bnu}) - \bar{A}^{(1)}_{\bnu}(\btheta) - K .
			\end{align*}
	\end{enumerate}
\end{lemma}

In Lemma \ref{lem:normalization}, Result \ref{lem:normalization:1} can be used to
build a function $f(\bnu)$ in order to
make the $A_{\bnu}$ independent from $\bnu$,
which is necessary to consider $A_{\bnu}$ as a $\log$-prior.
This change comes at a cost: the initial penalty $r^{(1)}_{\bnu}(\bmu)$
should be balanced by a non-constant term $f(\bnu)$.
Result \ref{lem:normalization:2} can be used to tune the constant $K$
in such a way that $\exp(A_{\bnu})$ integrates to $1$. 
As a result, $K$ is added to the penalty $r$, 
which does not influence the optimization process.
But $\exp(- \mathrm{Ent}(\beta_{0, \bnu}) - \bar{A}^{(1)}_{\bnu})$ 
must be integrable.

Also, it is worth noticing that, if Eqn.~\eqref{eqn:thm:1} 
admits a solution for $r_{\bnu}$, then it admits a solution
for $r_{\bnu} + K$, with any $K \in \mathbb{R}$.

\begin{corollary} \label{cor:main}
	We assume that the variational family $(\beta_{\bmu, \bnu})_{\bmu, \bnu}$
	fulfills Assumption~\ref{assum:1}.
	We define the following conditions on the penalty and the variational family:
	\begin{enumerate}[label=(\roman*)]
		\item $r_{\bnu} \in \mathcal{S}'(\mathbb{R}^N)$;
		\item the quotient $\nicefrac{\mathcal{F} r_{\bnu}}{\mathcal{F} \check{\beta}_{0, \bnu}}$
		belongs to $\mathcal{S}'(\mathbb{R}^N)$, where 
		$\check{\beta}_{0, \bnu}(\btheta) := \beta_{0, \bnu}(-\btheta)$;
		\item $A_{\bnu} := - \mathrm{Ent}(\beta_{0, \bnu})\, \mathbbm{1} \, - \, 
		\mathcal{F}^{-1}\left[ 	\frac{\mathcal{F} r_{\bnu}}{\mathcal{F} \check{\beta}_{0, \bnu}} \right]$ 
		does not depend on $\bnu$; let $A := A_{\bnu}$; \label{cond:d:indep}
		\item $A$ is such that $\int \exp (A) < \infty$. \label{cond:d':proper}
	\end{enumerate}
	
	If Conditions \ref{cond:c-:schwartz}-\ref{cond:d':proper} are fulfilled, then
	there exists a unique probability distribution $\alpha \in \mathcal{T}(\mathbb{R}^N)$
	such that:
	\begin{align}
	\exists K \in \mathbb{R} : \qquad
	& r_{\bnu}(\bmu) + K = \mathrm{KL}(\beta_{\bmu, \bnu} \| \alpha), \label{eqn:cor:1} \\
	\text{moreover:} \qquad
	& \alpha(\btheta) = \frac{\exp(A(\btheta))}{\int \exp(A(\mathbf{t})) \, \mathrm{d}\mathbf{t}} . \label{eqn:cor:2}
	\end{align}
	Incidentally, $K = \ln (\int \exp(A(\mathbf{t})) \, \mathrm{d}\mathbf{t})$.
	
	Conversely, if $\alpha \in \mathcal{T}(\mathbb{R}^N)$ is a probability distribution,
	and $K$ a constant, then the penalty $r_{\bnu}(\bmu) 
	:= \mathrm{KL}(\beta_{\bmu, \bnu} \| \alpha) - K$
	fulfills Conditions \ref{cond:c-:schwartz}-\ref{cond:d':proper}.
\end{corollary}


\subsection{Sum of Penalties} \label{sec:sum_penalties}

In the preceding section, we have established a result
in the general case of a penalty over the entire vector 
of means $\bmu$.
In practice, it is common \citep[Chap.\ 13]{shalev2014understanding} 
to deal with a penalty
that can be decomposed into a sum of sub-penalties, 
each one over a subset of the means $(\mu_1, \cdots, \mu_N)$.
Also, this framework encompasses
the ``weight decay'' technique \citep{krogh1992simple}, 
widely used in deep learning,
which consists in using a penalty proportional to the 
sum of the squared parameters of the model.
In this section, we show how such structure of the penalty
affects the structure of the corresponding prior.
In the simplest case when a penalty can be written as a sum
of $N$ penalties, each one over a component $\mu_i$ of $\bmu$,
then the corresponding prior $\alpha$ over $\btheta \in \mathbb{R}^N$ 
is a product of $N$ distributions:
$\alpha(\btheta) = \alpha_1(\theta_1) \cdots \alpha_N(\theta_N)$.

\begin{proposition} \label{prop:indep_pen}
	Let $(\beta_{\bmu, \bnu})_{\bmu, \bnu}$ a variational family fulfilling
	Assumption \ref{assum:1}, with parameters $\bmu \in \mathbb{R}^N$ and $\bnu$.
	Let $(I_i)_{1 \leq i \leq n}$ be a partition of $\{1, \cdots, N\}$.
	Given a vector $\btheta \in \mathbb{R}^N$, 
	we denote by $\btheta_{I_i}$ the vector of $\mathbb{R}^{|I_i|}$
	containing only the components of $\btheta$ whose index belongs to $I_i$.
	
	We consider the following penalty on $\beta_{\bmu, \bnu}$:
	\begin{align*}
		r_{\bnu}(\bmu) := \sum_{i = 1}^n r^{(i)}_{\bnu}(\bmu_{I_i}) .
	\end{align*}
	Thus, under Conditions \ref{cond:c-:schwartz} and \ref{cond:c:quotient}, 
	the solution of Eqn.~\eqref{eqn:thm:1} is:
	\begin{align}
		A_{\bnu}(\btheta) = - \WpEnt(\beta_{0, \bnu}) - \sum_{k = 1}^n 
		\mathcal{F}^{-1} \left[ \frac{\mathcal{F} r^{(i)}_{\bnu}}{\mathcal{F}
			\check{\beta}_{I_i | 0, \bnu}} \right](\btheta_{I_i}) , \label{eqn:indep:A}
	\end{align}
	where $\beta_{I_i | 0, \bnu}(\btheta_{I_i}) = \int \beta_{0, \bnu}(\btheta_{I_1}, \cdots, \btheta_{I_n}) 
	\, \mathrm{d}\btheta_{I_1} \cdots \mathrm{d}\btheta_{I_{i-1}} 
	\mathrm{d}\btheta_{I_{i+1}} \cdots \mathrm{d}\btheta_{I_n}$ is the marginal
	probability of $\beta_{0, \bnu}$ at indices $I_i$.
\end{proposition}

\begin{proof}
	According to Theorem \ref{thm:main}, the solution of Eqn.~\eqref{eqn:thm:1} is:
	\begin{align*}
		A_{\bnu} &= - \mathrm{Ent}(\beta_{0, \bnu})\, \mathbbm{1} \, - \, 
		\mathcal{F}^{-1}\left[ 	
		\frac{\mathcal{F} r_{\bnu}}{\mathcal{F} \check{\beta}_{0, \bnu}} \right] .
	\end{align*}
	
	We have
	$r_{\bnu}(\btheta) = \sum_{i = 1}^n r^{(i)}_{\bnu}(\btheta_{I_i}) \mathbbm{1}(\btheta)$,
	where $\mathbbm{1}$ is the constant function equal to $1$ on $\mathbb{R}^{N}$.
	So, $\mathcal{F}r_{\bnu} = \sum_{i = 1}^n (2 \pi)^{N - |I_i|} \mathcal{F} r^{(i)}_{\bnu} \, \delta|_{\bar{I}_i}$, 
	where $\delta|_{\bar{I}_i}$ is the Dirac distribution on functions of
	$(\btheta_{I_1}, \cdots, \btheta_{I_{i-1}}, \btheta_{I_{i+1}}, \cdots, \btheta_{I_n})$.
	
	Thus, we have:
	\begin{align*}
		\frac{\mathcal{F} r_{\bnu}}{\mathcal{F} \check{\beta}_{0, \bnu}}
		&= \sum_{i = 1}^n (2 \pi)^{N - |I_i|} \frac{\mathcal{F} r^{(i)}_{\bnu} \, 
			\delta|_{\bar{I}_i}}{\mathcal{F}\check{\beta}_{0, \bnu}}.
	\end{align*}
	
	Let us compute the right-hand side. Let $\varphi \in C^{\infty}_c(\mathbb{R}^N)$ be a test function. We have:
	\begin{align*}
		\left\langle \frac{\mathcal{F} r^{(i)}_{\bnu} \, 
			\delta|_{\bar{I}_i}}{\mathcal{F}\check{\beta}_{0, \bnu}}, \varphi \right\rangle 
		&= \left\langle \mathcal{F} r^{(i)}_{\bnu}, 
			\frac{\varphi(0, \cdots, 0, \cdot, 0, \cdots, 0)}%
			{(\mathcal{F}\check{\beta}_{0, \bnu})(0, \cdots, 0, \cdot, 0, \cdots, 0)} \right\rangle \\
		&= \left\langle \mathcal{F} r^{(i)}_{\bnu}, 
		\frac{\varphi(0, \cdots, 0, \cdot, 0, \cdots, 0)}%
		{\mathcal{F}\check{\beta}_{I_i|0, \bnu}} \right\rangle 
		= \left\langle \frac{\mathcal{F} r^{(i)}_{\bnu}}%
		{\mathcal{F}\check{\beta}_{I_i|0, \bnu}}\, \delta|_{\bar{I}_i}, \varphi \right\rangle .
	\end{align*}
	
	Therefore:
	\begin{align*}
		\mathcal{F}^{-1}\left[\frac{\mathcal{F} r_{\bnu}}{\mathcal{F} \check{\beta}_{0, \bnu}}\right]
		(\btheta)
		&= \sum_{i = 1}^n \mathcal{F}^{-1}\left[\frac{\mathcal{F} r^{(i)}_{\bnu}}%
		{\mathcal{F}\check{\beta}_{I_i|0, \bnu}}\right](\btheta_{I_i}) .
	\end{align*}
\end{proof}

\begin{remark} \label{rem:indep:uv}
	If $A_{\bnu}$ computed in Eqn.~\eqref{eqn:indep:A} is independent of $\bnu$, then 
	we can write each term of the sum in the following way:
	\begin{align*}
		f_i(\bnu, \btheta_{I_i}) := \mathcal{F}^{-1} \left[ \frac{\mathcal{F} r^{(i)}_{\bnu}}{\mathcal{F}
			\check{\beta}_{I_i | 0, \bnu}} \right](\btheta_{I_i})
		&= u_i(\btheta_{I_i}) + v_i(\bnu).
	\end{align*}
\end{remark}

\begin{proof}
	With simplified notation, we have:
	$A_{\bnu}(\btheta) = A(\btheta) = f_0(\bnu) + \sum_{i = 1}^{n} f_i(\bnu, \btheta_{I_i})$.
	Let $\bnu_0$ be some fixed value for the parameter $\bnu$. 
	Since $A_{\bnu}$ does not depend on $\bnu$, then, for any $\bnu$ and $\btheta$:
	\begin{align*}
		0 = A_{\bnu}(\btheta) - A_{\bnu_0}(\btheta) 
		= \sum_{i = 1}^{n} (f_i(\bnu, \btheta_{I_i}) - f_i(\bnu_0, \btheta_{I_i})).
	\end{align*}
	Thus, each term $f_i(\bnu, \btheta_{I_i}) - f_i(\bnu_0, \btheta_{I_i})$ is
	constant w.r.t.\ $\btheta_{I_i}$, so there exists $v_i$ s.t.:
	\begin{align*}
		f_i(\bnu, \btheta_{I_i}) = f_i(\bnu_0, \btheta_{I_i}) + v_i(\bnu).
	\end{align*}
	By setting $u_i(\btheta_{I_i}) = f_i(\bnu_0, \btheta_{I_i})$, the proof is achieved.
\end{proof}

\begin{corollary} \label{cor:indep_pen}
	With a penalty $r$ as defined in Proposition \ref{prop:indep_pen},
	under Conditions \ref{cond:c-:schwartz}-\ref{cond:d':proper}, 
	there exists a unique solution $\alpha \in \mathcal{T}(\mathbb{R}^N)$
	of Eqn.~\eqref{eqn:cor:1}:
	\begin{align*}
		\alpha(\btheta) &\propto \exp\left(- \WpEnt(\beta_{0, \bnu}) \right)
		\prod_{i = 1}^{n} \exp\left( -\mathcal{F}^{-1}\left[\frac{\mathcal{F} r^{(i)}_{\bnu}}%
		{\mathcal{F}\check{\beta}_{I_i|0, \bnu}}\right](\btheta_{I_i}) \right)
		= \prod_{i = 1}^n \alpha_i(\btheta_{I_i}) ,
	\end{align*}
	by using Remark \ref{rem:indep:uv}, where each $\alpha_i$ is a probability distribution
	over $\mathbb{R}^{I_i}$.
	The resulting $\alpha$ does not depend on $\bnu$ (Condition \ref{cond:d:indep}).
\end{corollary}

Under some conditions, Corollary \ref{cor:indep_pen} shows that, 
if a penalty $r_{\bnu}(\bmu) = \sum_{i = 1}^n r^{(i)}_{\bnu}(\bmu_{I_i})$
can be expressed as the influence of a Bayesian prior,
then this prior takes the form $\alpha(\btheta) = \prod_{i = 1}^n \alpha_i(\btheta_{I_i})$.
To summarize, a sum of penalties over different components
can be translated into a product of independent prior distributions over the same components:
\begin{align}
	\text{penalty } \quad r_{\bnu}(\bmu) = \sum_{i = 1}^n r^{(i)}_{\bnu}(\bmu_{I_i})
	\qquad \Rightarrow \qquad \text{ prior } \quad \alpha(\btheta) = \prod_{i = 1}^n \alpha_i(\btheta_{I_i}).
\end{align}

\begin{remark} \label{rem:relation_dep_posterior}
	Let $\beta_{\bmu, \bnu}(\btheta_{I_1}, \cdots, \btheta_{I_n})$ 
	be some probability distribution. 
	For a distribution $\alpha(\btheta) = \alpha_1(\btheta_{I_1}) \cdots \alpha_n(\btheta_{I_n})$,
	we have:
	\begin{align*}
		\mathrm{KL}(\beta_{\bmu, \bnu} \| \alpha)
		&= -\WpEnt(\beta_{\bmu, \bnu}) - \sum_{i = 1}^n 
		\int_{\mathbb{R}^{|I_i|}} \ln(\alpha_i(\btheta_{I_i})) \beta_{I_i | \bmu, \bnu}(\btheta_{I_i}) 
		\, \mathrm{d}\btheta_{I_i} \\
		&= \mathrm{KL}(\beta_{| \bmu, \bnu} \| \alpha) 
		+ \WpEnt(\beta_{| 0, \bnu}) -\WpEnt(\beta_{0, \bnu}) ,
	\end{align*}
	where $\beta_{| \bmu, \bnu}(\btheta) = \beta_{I_1 | \bmu, \bnu}(\btheta_{I_1}) \cdots 
	\beta_{I_n | \bmu, \bnu}(\btheta_{I_n})$ is the independent counterpart
	of $\beta_{\bmu, \bnu}(\btheta)$. 
	We have also used the fact that the entropy is independent of the mean of the distribution:
	for all $\bmu$, $\WpEnt(\beta_{| \bmu, \bnu}) = \WpEnt(\beta_{| 0, \bnu})$
	and $\WpEnt(\beta_{\bmu, \bnu}) = \WpEnt(\beta_{0, \bnu})$.
\end{remark}

\begin{example} \label{ex:bivariate}
	We consider the family of bivariate Gaussian distributions $(\beta_{\bmu, \bSigma})_{\bmu, \bSigma}$,
	with mean $\bmu \in \mathbb{R}^2$ and covariance matrix $\bSigma \in \mathbb{R}^{2 \times 2}$.
	We impose a penalty:
	\begin{align*}
		r_{\bSigma}(\bmu) := a(\bSigma) + \frac{b_1(\bSigma)}{2} \mu_1^2 + \frac{b_2(\bSigma)}{2} \mu_2^2,
	\end{align*}
	where $a$, $b_1$ and $b_2$ are functions.
	According to Proposition \ref{prop:indep_pen}, we have, with $s(x) = x^2$:
	\begin{align*}
		A_{\bSigma}(\btheta) = -\WpEnt(\beta_{0, \bSigma}) - a(\bSigma)
		- \frac{b_1(\bSigma)}{2} \mathcal{F}^{-1} \left[ \frac{\mathcal{F} s}{\mathcal{F}
			\check{\beta}_{1 | 0, \bSigma}} \right](\theta_1)
		- \frac{b_2(\bSigma)}{2} \mathcal{F}^{-1} \left[ \frac{\mathcal{F} s}{\mathcal{F}
			\check{\beta}_{2 | 0, \bSigma}} \right](\theta_2).
	\end{align*}
	
	We compute $\frac{\mathcal{F} s}{\mathcal{F} \check{\beta}_{1 | 0, \bSigma}}$.
	Let $\varphi \in C^{\infty}_c(\mathbb{R}^N)$ be a test function. We have:
	\begin{align*}
		\left\langle \frac{\mathcal{F} s}{\mathcal{F} \check{\beta}_{1 | 0, \bSigma}},
		\varphi \right\rangle 
		&= 2\pi \left\langle -\delta'',
		\varphi(t) \, \exp\left( \frac{\sigma_1^2 t^2}{2} \right) \right\rangle_t
		= 2\pi \left( -\varphi''(0) - \sigma_1^2\varphi(0) \right) \\
		&= 2 \pi \langle -\delta'' - \sigma_1^2 \delta, \varphi \rangle ,
	\end{align*}
	where $\sigma_i^2$ is the variance of the $i$-th marginal, $\theta_i \sim \beta_{i | 0, \bSigma}$.
	
	Since $\mathcal{F}^{-1}[-2 \pi \delta''] = s$ and $\mathcal{F}^{-1}[2 \pi \delta] = 1$, we have:
	\begin{align*}
		A_{\bSigma}(\btheta) &= -\WpEnt(\beta_{0, \bSigma}) - a(\bSigma)
		- \frac{b_1(\bSigma)}{2} (\theta_1^2 - \sigma_1^2) - \frac{b_2(\bSigma)}{2} (\theta_2^2 - \sigma_2^2) \\
		&= \left[-\WpEnt(\beta_{0, \bSigma}) - a(\bSigma) + \frac{b_1(\bSigma)}{2} \sigma_1^2
		+ \frac{b_2(\bSigma)}{2} \sigma_2^2\right] - \frac{b_1(\bSigma)}{2} \theta_1^2 - \frac{b_2(\bSigma)}{2} 
		\theta_2^2 .
	\end{align*}
	According to Condition \ref{cond:d:indep}, the function $A_{\bSigma}$ should not
	depend on $\bSigma$, that is:
	\begin{align*}
		b_1(\bSigma) &= b_1^0, &
		b_2(\bSigma) &= b_2^0, &
		a(\bSigma) &= a^0 -\WpEnt(\beta_{0, \bSigma}) + \frac{b_1^0}{2} \sigma_1^2
		+ \frac{b_2^0}{2} \sigma_2^2, 
	\end{align*}
	where $a^0$, $b_1^0$ and $b_2^0$ are constants. 
	Moreover, Condition \ref{cond:d':proper} imposes $b_1^0 > 0$ and $b_2^0 > 0$.
	Also, $\sigma_i^2 = \Sigma_{ii}$. So, if we denote by $|\bSigma|$ the determinant of $\bSigma$, we have:
	\begin{align*}
		r_{\bSigma}(\bmu) = a^0 - \frac{1}{2} \ln\left((2 \pi e)^2 |\bSigma|\right)
		+ \frac{b_1^0}{2} [\Sigma_{11} + \mu_1^2]
		+ \frac{b_2^0}{2} [\Sigma_{22} + \mu_2^2]  ,
	\end{align*}
	which is, up to a constant, the $\mathrm{KL}$ between $\mathcal{N}(\bmu, \bSigma)$
	and $\alpha = \mathcal{N}(0, 1/b_1^0) \otimes \mathcal{N}(0, 1/b_2^0)$.
\end{example}

According to Example \ref{ex:bivariate}, if the total penalty
can be split into two penalties, each one over a different parameter,
then the prior $\alpha$ is a product of independent univariate distributions.
Notably, this phenomenon occurs even if the variational family
contains multivariate distributions.

\section{Examples} \label{sec:examples}

In this section, we apply Theorem \ref{thm:main} and Corollary \ref{cor:main}
to a series of examples of variational families and penalties.
Our goal is to show how the initial choice of variational family and
penalty leads to a specific prior. Moreover,
we show that enforcing a Bayesian interpretation of the penalty
constrains the penalty itself: some penalties do not correspond to
any Bayesian prior.

In the following sections, we study several pairs of family/penalty.
Stated informally, the main results are:
\begin{itemize}
	\item if a $\mathcal{L}^2$ penalty over $\bmu$ corresponds to the influence
	of a Bayesian prior $\alpha$, then $\alpha$ is necessarily Gaussian
	(Sections \ref{sec:appl:univ_gaussian}, 
	\ref{sec:appl:l2_penalty}, \ref{sec:appl:multiv_gaussian});
	\item as a result, in that case, given a variational family $(\beta_{\bmu, \bnu})_{\bmu, \bnu}$, 
	the penalty consists in a competition between 
	$\WpEnt(\beta_{\bmu, \bnu})$, encouraged to be larger,
	and $\mathrm{Cov}(\beta_{\bmu, \bnu})$, encouraged to be smaller,
	balanced by the covariance of the Gaussian prior $\alpha$;
	\item when the penalty has a shape that is not fully 
	determined, e.g., $\beta_{\mu, \sigma} = \mathcal{N}(\mu, \sigma^2)$ and
	$r_{\sigma}(\mu) = a(\sigma) - b(\sigma) \cos(\frac{\mu}{\tau})$,
	with $a$ and $b$ undetermined functions, 
	we are able to propose strong constraints on $a$ and $b$
	so that $r$ can be interpreted as the influence of a Bayesian prior. 
	In this specific example, we obtain: 
	$a(\sigma) = a_0 - \frac{1}{2}\ln(2 \pi e \sigma^2)$, where $a_0$ is a constant, and 
	$b(\sigma) \propto \exp(- \frac{\sigma^2}{2 \tau^2})$ (Section \ref{sec:appl:cosine_penalty});
	\item the $\mathcal{L}^1$ penalty cannot be interpreted as
	the influence of a Bayesian prior if the variational family
	belongs to $\mathcal{S}(\mathbb{R})$, which includes
	the Gaussian family (see Section \ref{sec:appl:l1_penalty}).
\end{itemize}
We include several proofs to illustrate how Theorem \ref{thm:main}
and Corollary \ref{cor:main} can be applied.

\paragraph{Framework.}
Starting from here, we distinguish two kinds of variational families:
families of multivariate distributions \eqref{eqn:ex:multivar} and 
families of product of independent univariate distributions \eqref{eqn:ex:univar}.
For instance, if one looks for the posterior distribution of
a vector of parameters $\btheta \in \mathbb{R}^N$, 
within the family of Gaussian distributions,
one may consider two kinds of families:
\begin{align}
	\text{Multivariate: } \qquad \btheta &\sim \beta_{\bmu, \bSigma} 
	= \mathcal{N}(\bmu, \bSigma), \label{eqn:ex:multivar} \tag{M} \\
	\text{Prod.\ of Indep.\ Univariate: } \qquad \btheta &\sim \beta_{\bmu, \bsigma} 
	= \mathcal{N}(\mu_1, \sigma_1^2) \otimes \cdots \otimes
	\mathcal{N}(\mu_N, \sigma_N^2) , \label{eqn:ex:univar} \tag{PIU}
\end{align}
where $\bmu = (\mu_1, \cdots, \mu_N)$, $\bSigma \in \mathbb{R}^{N \times N}$,
and $\bsigma = (\sigma_1, \cdots, \sigma_N)$.

First, families \ref{eqn:ex:multivar} and \ref{eqn:ex:univar} are
very different from a practical point of view: in case 
\ref{eqn:ex:multivar}, the number of variational parameters is $N (N + 1)$,
while it is only $2 N$ in case \ref{eqn:ex:univar}.
So, it is easier and quicker to optimize $(\bmu, \bsigma)$ than $(\bmu, \bSigma)$
when $N$ is large, e.g., in neural networks.
Indeed, it is far more common to use \ref{eqn:ex:univar} when training neural networks 
with VI \citep{graves2011practical}.

Second, even when using a \ref{eqn:ex:multivar} variational family 
$(\beta_{\bmu, \bnu})_{\bmu, \bnu}$ in combination with a penalty
$r_{\bnu}(\bmu) = r^{(0)}(\bnu) + \sum_{i = 1}^N r^{(i)}_{\bnu}(\mu_i)$,
Corollary \ref{cor:indep_pen} ensures that the corresponding prior, 
if it exists, takes the form: $\alpha(\btheta) = \alpha_1(\theta_1) \cdots \alpha_N(\theta_N)$.
Thus, we can use Corollary \ref{cor:indep_pen}
and Remark \ref{rem:relation_dep_posterior}:
first, we look for the $(\alpha_i)_i$ solving:
\begin{align*}
	r^{(i)}_{\bnu}(\mu_i) = \mathrm{KL}(\beta_{i| \bmu, \bnu} \| \alpha_i) ,
\end{align*}
then we compute $\alpha(\btheta) = \alpha_1(\theta_1) \cdots \alpha_N(\theta_N)$.
Finally, we recover the penalty:
\begin{align*}
	r_{\bnu}(\bmu) = \mathrm{KL}(\beta_{| \bmu, \bnu} \| \alpha) 
	+ \WpEnt(\beta_{| 0, \bnu}) - \WpEnt(\beta_{0, \bnu})
	= \mathrm{KL}(\beta_{\bmu, \bnu} \| \alpha) .
\end{align*}

Therefore, we focus mainly on cases where both priors and
candidate posteriors are assumed to be \ref{eqn:ex:univar}:
they cover a wide range of practical cases,
and also cases with \ref{eqn:ex:multivar} candidate posteriors
and \ref{eqn:ex:univar} priors, which reduce to \ref{eqn:ex:univar}
priors and posteriors.

\subsection[\texorpdfstring{Univariate Gaussian with $\mathcal{L}^2$ Penalty}%
		{Univariate Gaussian with L2 Penalty}]
	{Univariate Gaussian with $\mathcal{L}^2$ Penalty}  \label{sec:appl:univ_gaussian}

Let us study the family of univariate Gaussian distributions (Example~\ref{ex:gaussian}) 
with an $\mathcal{L}^2$ penalty.
Formally:
\begin{align*}
	\beta_{\mu, \sigma}(\theta) &= \frac{1}{\sqrt{2 \pi \sigma^2}} \exp\left( - \frac{(\theta - \mu)^2}{2 \sigma^2} 
	\right), &
	r_{\sigma}(\mu) &= a(\sigma) + \frac{1}{2} b(\sigma) \mu^2 ,
\end{align*}
where $a$ and $b$ are two functions.

\begin{corollary}\label{cor:appl:univ_gaussian}
	If the penalty $r_{\sigma}$ above corresponds to a prior $\alpha$, then
	$\alpha$ is a Gaussian $\mathcal{N}(0, \sigma_0^2)$.
	Moreover, $a$ and $b$ must have the following form:
	\begin{align*}
		a(\sigma) &= \frac{1}{2} \left[\frac{\sigma^2}{\sigma_0^2} 
		+ \ln\left(\frac{\sigma_0^2}{\sigma^2}\right) - 1\right],
		& b(\sigma) &= \frac{1}{\sigma_0^2} .
	\end{align*}
\end{corollary}

\begin{proof}
	This result is an application of Corollary \ref{cor:main}.
	The conditions \ref{cond:a-:translation}-\ref{cond:a:schwartz}
	and \ref{cond:c-:schwartz}-\ref{cond:c:quotient} of Theorem \ref{thm:main} are fulfilled.
	Thus, by using \eqref{eqn:thm:2}: 
	\begin{align*}
		A_{\sigma}(\theta) = - \frac{1}{2}\ln(2 \pi e \sigma^2) - \mathcal{F}^{-1}\left[
		\frac{\mathcal{F}r_{\sigma}}{\mathcal{F} \check{\beta}_{0, \sigma}}\right] .
	\end{align*}
	Let $s(\theta) = \theta^2$. 
	We have, for any test function $\varphi \in C^{\infty}_c(\mathbb{R})$:
	\begin{align*}
		\left\langle \frac{\mathcal{F}s}{\mathcal{F} \check{\beta}_{0, \sigma}},
		\varphi \right\rangle
		&= - 2 \pi \left\langle \frac{\delta''}{\exp\left(-\frac{\sigma^2 t^2}{2}\right)},
		\varphi(t) \right\rangle_t
		= - 2 \pi \left\langle \delta'',
		\varphi(t) \exp\left(\frac{\sigma^2 t^2}{2}\right) \right\rangle_t \\
		&= -2 \pi (\varphi''(0) + \sigma^2 \varphi(0))
		= -2 \pi \langle \delta'' + \sigma^2 \delta, \varphi \rangle .
	\end{align*}
	Thus: $A_{\sigma}(\theta) = - \frac{1}{2}\ln(2 \pi e \sigma^2) - a(\sigma) - \frac{1}{2} b(\sigma)
		(\theta^2 - \sigma^2)$,
	which is a constant function of $\theta$ (Condition \ref{cond:d:indep}) if,
	and only if, there exists two constants $a_0$ and $b_0$ such that:
	\begin{align*}
		a(\sigma) &= - \frac{1}{2} \ln(2 \pi e \sigma^2) + \frac{1}{2} b(\sigma) \sigma^2 - a_0, 
		& b(\sigma) &= b_0.
	\end{align*}
	So, $\alpha = \exp(A)$ is a density of probability (Condition \ref{cond:d':proper})
	if $b_0 > 0$ and $\int \alpha = 1$.
	If so, $\alpha$ is the density of a Gaussian $\mathcal{N}(0, \sigma_0^2)$,
	with $b_0 = \frac{1}{\sigma_0^2}$ and 
	$a_0 = -\frac{1}{2} \ln(2 \pi \sigma_0^2)$.
\end{proof}

Thus, the assumption that a penalty $r$
arises from a variational interpretation
is a strong constraint over $r$. Here, the 
penalty $r$ was initially parameterized by a pair of real functions $(a, b)$, and is finally parameterized by 
a single number $\sigma_0^2$.


\subsection{Dirac Family and the MAP}
\label{sec:appl:dirac}

Another basic example
is to use Dirac distributions as variational posteriors
(Example~\ref{ex:dirac}): $\beta_{\mu, \bnu} = \delta_\mu$. 
Since $\delta_0 \notin \mathcal{S}(\mathbb{R})$, Condition \ref{cond:a:schwartz} 
is not satisfied. However, it is possible to apply 
Formula~\eqref{eqn:thm:2} and check that the resulting prior $\alpha$ is
consistent with a chosen penalty $r$.
Applying Formula~\eqref{eqn:thm:2} yields
\begin{align*}
A &= - \mathrm{Ent}(\delta_0) \mathbbm{1} - \mathcal{F}^{-1}\left[ 
\frac{\mathcal{F} r}{\mathcal{F} \delta_0} \right] 
= - \mathcal{F}^{-1}\left[ \frac{\mathcal{F} r}{\mathbbm{1}} \right] = - r .
\end{align*}

Thus, if $\exp (-r)$ integrates to $0<\kappa<\infty$, then we can define $\alpha =
\frac{1}{\kappa} \exp (-r)$. Then, we can check that
indeed $\mathrm{KL}(\delta_{\mu} \| \alpha) = r(\mu)$ up to a constant:
\begin{align*}
\mathrm{KL}(\delta_{\mu} \| \alpha) &= -\WpEnt(\delta_0) - \left\langle
\delta_{\mu} , \ln (\alpha)
\right\rangle \\
&= - \ln \alpha(\mu)-\WpEnt(\delta_0) \\
&= r(\mu) + \ln \kappa -\WpEnt(\delta_0),
\end{align*}
which confirms that the proposed prior $\alpha$ is consistent with the
penalty $r(\cdot)$.

Thus, this formula recovers via variational inference the well-known
penalty--prior equivalence in the MAP approximation,
$\alpha_{\mathrm{VI}}(\theta) \propto \exp(-
r(\theta)) \propto \alpha_{\mathrm{MAP}}(\theta)$
\citep{mackay1992practical}.

However, this is somewhat formal:
the entropy $\WpEnt(\delta_0)$ of a Dirac function is technically undefined and is an ``infinite
constant''. In practice, though, with a finite machine precision
$\epsilon$,
a Dirac mass can be defined as a uniform distribution over
an interval of size $\epsilon$, and $\WpEnt(\delta_0)$ becomes the finite
constant $\ln \epsilon$.

\subsection[\texorpdfstring{Arbitrary Variational Family with $\mathcal{L}^2$ Penalty}%
{Arbitrary Variational Family with L2 Penalty}]
{Arbitrary Variational Family with $\mathcal{L}^2$ Penalty} \label{sec:appl:l2_penalty}

Let us consider a family of univariate variational posteriors $(\beta_{\mu, \bnu})_{\mu, \bnu}$ and a penalty:
\begin{align}
	r_{\bnu}(\mu) := a(\bnu) + \frac{1}{2} b(\bnu) \mu^2 , \label{eqn:appl:l2_penalty:pen}
\end{align}
where $a$ and $b$ are functions. 
We assume that all Conditions \ref{cond:a-:translation}-\ref{cond:b:nonzero}
and \ref{cond:c-:schwartz}-\ref{cond:d':proper} are fulfilled.

\begin{corollary} \label{cor:appl:l2_penalty}
	If $r$ corresponds to a $\mathrm{KL}$-divergence with prior $\alpha$, 
	then there exists $\sigma_0 > 0$ s.t.:
	\begin{align}
		r_{\bnu}(\mu) &= \frac{\mathrm{Var}(\beta_{0, \bnu}) + \mu^2}{2 \sigma_0^2} 
		- \WpEnt(\beta_{0, \bnu}) + \frac{1}{2}\ln(2 \pi e \sigma_0^2), \label{eqn:appl:l2_penalty:cor:1} \\
		\alpha(\theta) &= \frac{1}{\sqrt{2 \pi \sigma_0^2}} 
		\exp\left(- \frac{\mu^2}{2 \sigma_0^2}\right) . \nonumber
	\end{align}
\end{corollary}

\begin{proof}
	Formula \eqref{eqn:thm:2} yields: 
	\begin{align*}
	A_{\bnu}(\theta) = - \WpEnt(\beta_{0, \bnu}) - \mathcal{F}^{-1}\left[
	\frac{\mathcal{F}r_{\bnu}}{\mathcal{F} \check{\beta}_{0, \bnu}}\right] .
	\end{align*}
	Let $s(\theta) = \theta^2$. 
	We have, for any test function $\varphi \in C^{\infty}_c(\mathbb{R})$:
	\begin{align*}
	\left\langle \frac{\mathcal{F}s}{\mathcal{F} \check{\beta}_{0, \bnu}},
	\varphi \right\rangle
	&= - 2 \pi \left\langle \delta'',
	\frac{\varphi(t)}{\check{\beta}_{0, \bnu}(t)}  \right\rangle_t
	= - 2 \pi [\varphi(0) + \mathrm{Var}(\check{\beta}_{0, \bnu}) \varphi''(0)],
	\end{align*}
	since $(\mathcal{F} \check{\beta}_{0, \bnu})(0) = 1$,
	$(\mathcal{F} \check{\beta}_{0, \bnu})'(0) = 0$ ($\check{\beta}_{0, \bnu}$ is centered)
	and $(\mathcal{F} \check{\beta}_{0, \bnu})''(0) = \mathrm{Var}(\check{\beta}_{0, \bnu})$.
	
	Thus: $A_{\bnu}(\theta) = - \mathrm{Ent}(\beta_{0, \bnu}) - a(\bnu) 
	- \frac{1}{2} b(\bnu) (\theta^2 - \mathrm{Var}(\beta_{0, \bnu}))$,
	which is a constant function of $\theta$ (Condition \ref{cond:d:indep}) if,
	and only if, there exists two constants $a_0$ and $b_0$ such that:
	\begin{align*}
	a(\bnu) &= - \mathrm{Ent}(\beta_{0, \bnu}) + \frac{1}{2} b(\bnu) \mathrm{Var}(\beta_{0, \bnu}) - a_0, 
	& b(\bnu) &= b_0.
	\end{align*}
	So, $\alpha = \exp(A)$ is a density of probability (Condition \ref{cond:d':proper})
	if $b_0 > 0$ and $\int \alpha = 1$.
	If so, $\alpha$ is the density of a Gaussian $\mathcal{N}(0, \sigma_0^2)$,
	with $b_0 = \frac{1}{\sigma_0^2}$ and 
	$a_0 = -\frac{1}{2} \ln(2 \pi \sigma_0^2)$.
\end{proof}

So, in general, if we penalize $\mu$ with a term $\mu^2$ and we are in a Bayesian setup, 
then the corresponding prior is necessarily Gaussian.

\begin{remark}
	Up to a constant, $r_{\bnu}$ is:
	\begin{align*}
		r_{\bnu}(\mu) &= \left[\frac{1}{2}\frac{\mathrm{Var}(\beta_{0, \bnu})}{\mathrm{Var}(\alpha)} 
		- \left(\WpEnt(\beta_{0, \bnu}) - \WpEnt(\alpha)\right) \right]
		+ \frac{\mu^2}{2 \sigma_0^2} .
	\end{align*}
	So, there is a competition between the entropy term, which encourages the distribution 
	$\beta_{\mu, \bnu}$ to be flatter, and the variance term, which pushes its variance
	towards $0$. These two terms are balanced by the effect of the prior $\alpha$, through its
	own variance $\mathrm{Var}(\alpha) = \sigma_0^2$.
\end{remark}

In this setup, the shape of the prior is entirely determined by the penalty,
which means that the choice of the variational family does not matter.
So, this is a generalization of Corollary \ref{cor:appl:univ_gaussian} 
obtained in Section \ref{sec:appl:univ_gaussian} for a Gaussian variational family.
This striking result is a consequence of our choice for the family of penalties.
Indeed, our initial penalty $r$ \eqref{eqn:appl:l2_penalty:pen} 
is parameterized by two functions $a$ and $b$, and we have discovered that,
if $r$ corresponds to a $\mathrm{KL}$-divergence, then
$a$ and $b$ are constrained to a specific shape (see Eqn.~\eqref{eqn:appl:l2_penalty:cor:1}).
Additionally, $a$ and $b$ are not direct functions of $\bnu$,
but rather functions of $\WpEnt(\beta_{0, \bnu})$ 
and $\mathrm{Var}(\beta_{0, \bnu})$.

\begin{remark}
	Moreover, the Fourier transform of the penalty $r(x) = a + b x^2$,
	is a linear combination of $\delta$
	and $\delta''$, both with support $\{0\}$.
	Due to the relation between the $k$-th derivatives of $\mathcal{F} \beta$ at $0$
	and the $k$-th moments of the distribution $\beta$,
	the variance of $\beta$ arises naturally in the constrained penalty $r$
	(see Eqn.~\eqref{eqn:appl:l2_penalty:cor:1}).
	So, if we consider $\mathcal{L}^{2 p}$ penalties with $p > 0$ integer, 
	$p$-th moments should arise as well in the constrained penalty. 
\end{remark}

\subsection[\texorpdfstring{Multivariate Gaussian with $\mathcal{L}^2$ Penalty}%
{Multivariate Gaussian with L2 Penalty}]
{Multivariate Gaussian with $\mathcal{L}^2$ Penalty}  \label{sec:appl:multiv_gaussian}

Let us study the family of multivariate Gaussian distributions with a $\mathcal{L}^2$ penalty.
Formally:
\begin{align*}
\beta_{\bmu, \bSigma}(\btheta) &= \frac{1}{\sqrt{(2 \pi)^N |\bSigma|}} 
\exp\left( - \frac{(\btheta - \bmu)^T \bSigma^{-1} (\btheta - \bmu)}{2} \right), \\
r_{\bSigma}(\bmu) &= a(\bSigma) + \frac{1}{2} \bmu^T B(\bSigma) \bmu ,
\end{align*}
where $a : \mathbb{R}^{N \times N} \rightarrow \mathbb{R}$ 
and $B : \mathbb{R}^{N \times N} \rightarrow \mathbb{R}^{N \times N}$ are two functions.

\begin{corollary}\label{cor:appl:multiv_gaussian}
	If the penalty $r_{\bSigma}$ above corresponds to a prior $\alpha$, then
	$\alpha$ is a Gaussian $\mathcal{N}(0, \mathbf{B}_0^{-1})$.
	Moreover, $a$ and $B$ must have the following form:
	\begin{align*}
	a(\bSigma) &= \frac{1}{2} \Big[\mathrm{Tr}(\mathbf{B}_0 \bSigma) 
	- \ln\left(|\mathbf{B}_0| \, |\bSigma| \right) - N\Big],
	& B(\bSigma) &= \mathbf{B}_0 .
	\end{align*}
\end{corollary}

\begin{proof}
	This result is an application of Corollary \ref{cor:main}.
	Conditions \ref{cond:a-:translation}-\ref{cond:b:nonzero}
	and \ref{cond:c-:schwartz}-\ref{cond:c:quotient}
	of Theorem \ref{thm:main} are fulfilled.
	Thus, by using \eqref{eqn:thm:2}: 
	\begin{align*}
	A_{\bSigma}(\btheta) = - \frac{1}{2}\ln\left((2 \pi e)^N |\bSigma|\right) - \mathcal{F}^{-1}\left[
	\frac{\mathcal{F}r_{\bSigma}}{\mathcal{F} \check{\beta}_{0, \bSigma}}\right] .
	\end{align*}
	Let $s(\btheta) = \btheta^T B(\bSigma) \btheta$. 
	We have, for any test function $\varphi \in C^{\infty}_c(\mathbb{R})$:
	\begin{align*}
	\left\langle \frac{\mathcal{F}s}{\mathcal{F} \check{\beta}_{0, \sigma}},
	\varphi \right\rangle
	&= \left\langle \mathcal{F}\left[ \sum_{i, j = 1}^{N} B(\bSigma)_{ij} \theta_i \theta_j \right],
	\varphi(\mathbf{t}) \exp\left(\frac{\mathbf{t}^T\bSigma \mathbf{t}}{2}\right) \right\rangle_{\mathbf{t}} \\
	&= - (2 \pi)^N \sum_{i, j = 1}^{N} B(\bSigma)_{ij} \left\langle \partial^{p_{ij}} \delta,
	\varphi(\mathbf{t}) \exp\left(\frac{\mathbf{t}^T\bSigma \mathbf{t}}{2}\right) \right\rangle_{\mathbf{t}},
	\end{align*}
	where $\partial^{p_{ij}} \delta : \varphi \mapsto \frac{\partial^2 \varphi}{\partial t_i \partial t_j}(0)$.
	So:
	\begin{align*}
	\left\langle \frac{\mathcal{F}s}{\mathcal{F} \check{\beta}_{0, \sigma}},
	\varphi \right\rangle
	&= - (2 \pi)^N \sum_{i, j = 1}^{N} \left[B(\bSigma)_{ij} \Sigma_{ij} \left\langle \delta,
	\varphi \right\rangle + B(\bSigma)_{ij} \left\langle \partial^{p_{ij}} \delta,
	\varphi \right\rangle\right] \\
	&= - (2 \pi)^N \left\langle \mathrm{Tr}(B(\bSigma) \bSigma) \delta 
	+ \sum_{i,j = 1}^N B(\bSigma)_{ij} \partial^{p_{ij}} \delta, \varphi \right\rangle .
	\end{align*}
	Thus:
	\begin{align*}
	\mathcal{F}^{-1} \left[\frac{\mathcal{F}s}{\mathcal{F} \check{\beta}_{0, \sigma}}\right](\btheta)
	&= \btheta^T B(\bSigma) \btheta - \mathrm{Tr}(B(\bSigma) \bSigma) \\
	A_{\bSigma}(\btheta) &= - \frac{1}{2} \ln\left((2 \pi e)^N |\bSigma|\right) - a(\bSigma)
	+ \frac{1}{2} \mathrm{Tr}(B(\bSigma) \bSigma) - \frac{1}{2} \btheta^T B(\bSigma) \btheta ,
	\end{align*}
	where $A_{\bSigma}(\theta)$ is a constant function of $\theta$ (Condition \ref{cond:d:indep}) if,
	and only if, there exists two constants $a_0 \in \mathbb{R}$ 
	and $\mathbf{B}_0 \in \mathbb{R}^{N \times N}$ such that:
	\begin{align*}
	a(\bSigma) &= - \frac{1}{2} \ln\left((2 \pi e)^N |\bSigma|\right) 
	- a_0 + \frac{1}{2} \mathrm{Tr}(B(\bSigma) \bSigma), 
	& B(\bSigma) &= \mathbf{B}_0.
	\end{align*}
	So, $\alpha = \exp(A)$ is a density of probability (Condition \ref{cond:d':proper})
	if $\mathbf{B}_0$ is positive definite and $\int \alpha = 1$.
	If so, $\alpha$ is the density of a Gaussian $\mathcal{N}(0, \mathbf{B}_0^{-1})$,
	$a_0 = -\frac{1}{2} \ln((2 \pi)^N |\mathbf{B}_0^{-1}|)$.
\end{proof}

\begin{remark}
	More generally, one can deal with any multivariate family $(\beta_{\bmu, \bnu})_{\bmu, \bnu}$,
	and not only the Gaussian, if the conditions \ref{cond:a-:translation}-\ref{cond:b:nonzero}
	and \ref{cond:c-:schwartz}-\ref{cond:d':proper} are fulfilled.
	Then, one can easily adapt the proof of Corollary \ref{cor:appl:l2_penalty}.
	With $r_{\bnu}(\bmu) = a(\bnu) + \frac{1}{2} \bmu^T B(\bnu) \bmu$ ,
	we have:
	\begin{align*}
	r_{\bnu}(\bmu) &= \frac{1}{2} \bmu^T \mathbf{B}_0 \bmu 
	+ \mathrm{Tr}(\mathbf{B}_0 \mathrm{Cov}(\beta_{0, \bnu}))
	+\frac{1}{2} \ln\left((2 \pi e)^N |\mathbf{B}_0^{-1}|\right)
	- \WpEnt(\beta_{0, \bnu}) \\
	\alpha(\btheta) &= (2 \pi)^{-N/2} |\mathbf{B}_0|^{1/2}
	\exp\left(- \frac{1}{2} \btheta^T \mathbf{B}_0 \btheta\right) . 
	\end{align*}
	So, even in the multivariate case, under some technical conditions, 
	the $\mathcal{L}^2$ penalty corresponds always to a 
	(multivariate) Gaussian prior.
	Again, the parameters $\bnu$ of $\beta_{\bmu, \bnu}$ are
	constrained through a competition between the entropy of $\beta_{\bmu, \bnu}$
	(which should be large) and its covariance (which should be small).
	This competition is modulated by the covariance $\mathbf{B}_0^{-1}$ of the prior $\alpha$.
\end{remark}

\subsection{Univariate Gaussian with Cosine Penalty} \label{sec:appl:cosine_penalty}

Let $(\beta_{\mu, \sigma})_{\mu, \sigma}$ be the family of 
univariate Gaussian distributions $\beta_{\mu, \sigma} = \mathcal{N}(\mu, \sigma^2)$.
In a context of quantization, a periodic penalty, such as cosine%
\footnote{\cite{naumov2018periodic} use $\sin^2$ instead of $\cos$.},
can be used \citep{naumov2018periodic}.
Such a penalty would encourage the parameters of the model to remain close to
values that can be encoded with few bits:
\begin{align*}
r(\beta_{\mu, \sigma^2}) := a(\sigma^2) - b(\sigma^2) \cos\left(\frac{\mu}{\tau}\right) ,
\end{align*}
where $\tau > 0$. With such penalty, if $b(\sigma^2) > 0$, $\mu$ is
pushed towards the points $\{ 2 k \tau \pi : k \in \mathbb{Z}\}$.
We show that, if this penalty corresponds to a prior (possibly improper),
then the functions $a$ and $b$ must have a specific shape
with properties we detail below.

\begin{corollary}
	Let us consider the Gaussian variational family with penalty:
	\begin{align*}
	r(\beta_{\mu, \sigma^2}) := a(\sigma^2) - b(\sigma^2) \cos\left(\frac{\mu}{\tau}\right) ,
	\end{align*}
	where $\tau > 0$.
	Thus, if $r$ corresponds to a $\mathrm{KL}$-divergence with improper prior $\alpha$, 
	then there exist $a_0, b_0 \in \mathbb{R}$ s.t.:
	\begin{align*}
	\alpha(\theta) &\propto \exp\left(b_0 \cos\left(\frac{\mu}{\tau}\right)\right), \\
	r(\beta_{\mu, \sigma^2}) &= a_0 - \frac{\ln(2 \pi e \sigma^2)}{2} 
	- b_0 e^{-\frac{\sigma^2}{2 \tau^2}} \cos\left(\frac{\mu}{\tau}\right) .
	\end{align*}
\end{corollary}

Let us assume that $b_0 > 0$.
In this setup, $\mu$ is pushed towards the points $\{ 2 k \tau \pi : k \in \mathbb{Z}\}$
with strength $\exp(-\frac{\sigma^2}{2 \tau^2})$.
So, if $\sigma \gg \tau$, then $\mu$ is very weakly penalized.
On the contrary, if $\sigma \ll \tau$, $\mu$ is penalized with strength $b_0$.
In short, if the standard deviation $\sigma$ of $\beta_{\mu, \sigma^2}$ is
much larger that the width $\tau$ of the oscillations of the penalty,
then $\mu$ is not penalized. 

This behavior is consistent
with what we expect when we train $(\mu, \sigma)$:
if $\sigma$ is so large that the center of the distribution $\beta_{\mu, \sigma}$
covers several oscillations of the penalty, 
then the samples $\theta \sim \beta_{\mu, \sigma}$
would lead to values of $\cos(\frac{\theta}{\tau})$
that are independent of $\mu$.
So, there would be no point in optimizing $\mu$.
On the contrary, if $\sigma \ll \tau$, then almost all samples $\theta \sim \beta_{\mu, \sigma}$
would lead to $\cos(\frac{\theta}{\tau}) \approx \cos(\frac{\mu}{\tau})$.
So, optimizing $\mu$ would be meaningful to obtain better samples.

From the point of view of $\sigma$, if the penalty $- \cos(\frac{\mu}{\tau})$
is large, then smaller $\exp(-\frac{\sigma^2}{2 \tau^2})$ are encouraged,  
which means larger $\sigma$. On the contrary, if $- \cos(\frac{\mu}{\tau})$
is already negative, then $\sigma$ is encouraged to be smaller.
To summarize, if $\mu$ is strongly penalized, then $\sigma$
tends to increase, making the samples $\theta \sim \beta_{\mu, \sigma}$ explore more,
and the more $\mu$ is close to a minimum, the more the samples 
exploit the region around $\mu$.

\subsection[\texorpdfstring{Smooth Variational Family with $\mathcal{L}^1$ Penalty}%
{Smooth Variational Family with L1 Penalty}]
{Smooth Variational Family with $\mathcal{L}^1$ Penalty} \label{sec:appl:l1_penalty}

In Section \ref{sec:penalty_KL}, we have proven results
if some restrictive conditions are fulfilled.
Specifically, for a variational family $(\beta_{\mu, \bnu})_{\mu, \bnu}$,
we need $r_{\bnu} \in \mathcal{S}'(\mathbb{R})$ in order to be able to compute
$r_{\bnu}$ (see Theorem \ref{thm:main}).
Since $r : x \mapsto |x|$ does not belong to $\mathcal{S}'(\mathbb{R})$,
then we cannot apply Theorem \ref{thm:main} to the $\mathcal{L}^1$
penalty, which is quite common.
So, one could argue that the conditions on $r$ are too restrictive .

However, according to Proposition \ref{cor:penalty_L1} below,
when the distributions $\beta_{\mu, \bnu}$ of the variational
family are smooth enough, it is impossible to find a
prior $\alpha$ s.t.:
\begin{align*}
	\exists K \in \mathbb{R} : \forall \mu \in \mathbb{R}, \quad |\mu| + K 
	= \mathrm{KL}(\beta_{\mu, \bnu} \| \alpha).
\end{align*}

\begin{proposition} \label{cor:penalty_L1}
	If the variational family $(\beta_{\mu, \bnu})_{\mu, \bnu}$ 
	is included in $\mathcal{S}(\mathbb{R})$
	and the penalty $r$ does not belong to $C^{\infty}(\mathbb{R})$, 
	then it is impossible to find any prior $\alpha \in \mathcal{T}(\mathbb{R})$ 
	(even improper) such that:
	\begin{align*}
		\exists K \in \mathbb{R} : \quad 
		r_{\bnu}(\mu) + K = \mathrm{KL}(\beta_{\mu, \bnu} \| \alpha) .
	\end{align*}
\end{proposition}

Proposition \ref{cor:penalty_L1} is an immediate consequence of a
property of the convolution:
\begin{proposition}{\citep[Chap.\ I, Theorem 3.13]{stein2016introduction}} \label{prop:convolution_smooth}
	Let $A \in \mathcal{S}'(\mathbb{R})$ and $\beta_{0, \bnu} \in \mathcal{S}(\mathbb{R})$.
	Thus, $A * \beta_{0, \bnu} \in C^{\infty}(\mathbb{R})$.
\end{proposition}

This negative result is corroborated by \cite{kaban2007bayesian},
who pointed out that a Laplace prior
is sparsifying in the MAP case, but not when computing the full Bayesian posterior.
The smoothing effect present in Proposition \ref{prop:convolution_smooth}
has also been noticed.
So, the ``prior'' $\alpha(x) \propto \exp(-|x|)$ corresponding 
to the penalty $r(x) = |x|$ in the MAP case
is no longer valid, and cannot even be tuned to generate a 
$\mathrm{KL}$ proportional to $|\mu|$, when considering 
smooth candidate posteriors.

\begin{remark}
	This regularity condition on $r$ is implicitly contained in Condition \ref{cond:c:quotient}:
	$T := \nicefrac{\mathcal{F} r_{\bnu}}{\mathcal{F} \check{\beta}_{0, \bnu}} \in \mathcal{S}'(\mathbb{R}^N)$.
	In other words, if $r_{\bnu}$ is not regular enough compared to $\check{\beta}_{0, \bnu}$,
	$T$ would ``diverge too quickly to infinity'', so $T \notin \mathcal{S}'(\mathbb{R}^N)$.
\end{remark}

\section{Building a Heuristic for the Penalty Factor} \label{sec:application}

We show how our main results can be used to provide a heuristic
on the penalty factor and the variance of the prior in the context of 
neural networks. Specifically, this heuristic leads to
a non-isotropic diagonal prior, which contrasts with commonly used priors
in Bayesian neural networks. 
On a small set of experiments, this heuristic provides a penalty factor 
relatively close to the optimal one.
Additionally, we discuss the relation between our results and the
results obtained in the framework of the ``cold posterior effect''.

\subsection{Tuning the Penalty Factor}

Let $\mathcal{B} = \{\beta_{\mathbf{u}}\}$ be a variational family. 
Given a custom penalty $r$ and a \emph{penalty factor} $\lambda > 0$, we aim to minimize 
the following loss on $\mathcal{B}$:
\begin{align}
\mathcal{L}(\beta_{\mathbf{u}}) := - \mathbb{E}_{\btheta \sim \beta_{\mathbf{u}}} \ln
p_{\btheta}(\mathcal{D}) + \lambda r(\mathbf{u}) . \label{eqn:loss}
\end{align}

We first have to tune the hyperparameter $\lambda$, which balances the
amount of regularization we want to set when training the model. Usually, $\lambda$ 
is determined by performing cross-validation \citep{stone1974cross}.
In the specific case of neural networks, cross-validation necessitates 
as many training runs as the number of different $\lambda$ to be tested.

In this document, we provide a heuristic designed to output a reasonable value for $\lambda$.
More specifically, this heuristic is made of three bricks:
\begin{enumerate}[label=\arabic*.]
\item interpret the penalty $\lambda r(\mathbf{u})$ as the influence of 
a Bayesian prior $\alpha_{\lambda}$;
\item assume that the prior $\alpha_{\lambda}$ should have the same variance as a
reasonable initialization distribution of the weights, e.g., the one proposed by 
\cite{glorot2010understanding}, denoted by $\alpha^{\star}$;
\item we set $\lambda$ in such a way that: 
$\mathrm{Var}(\alpha_{\lambda}) = \mathrm{Var}(\alpha^{\star})$.
\end{enumerate}
Then, we test this heuristic with different neural networks.

\subsection{Building the heuristic}

We illustrate our method in the case of a Gaussian variational family
and a $\mathcal{L}^2$ penalty over the means:
\begin{align*}
\beta_{\mathbf{u}} &:= \beta_{\bmu, \bsigma} = 
\mathcal{N}(\mu_1, \sigma_1^2) \otimes \cdots \otimes \mathcal{N}(\mu_N, \sigma_N^2) , \\
r_{\bsigma}(\bmu) &:= \sum_{i = 1}^N r_{i,\bsigma}(\mu_i)
= \sum_{i = 1}^{N} \left[ a_i(\bsigma) + b_i(\bsigma) \mu_i^2 \right] .
\end{align*}

\paragraph{Step 1: interpretation of the penalty.}
In this framework, the requirements of Corollary 2 are fulfilled, and
the prior $\alpha$ corresponding to $r$ is:
\begin{align*}
\alpha(\btheta) := \alpha_1(\theta_1) \cdots \alpha_N(\theta_N) ,
\end{align*}
where each $\alpha_i$ is a Gaussian distribution with mean $0$.

From now, and without loss of generality, we focus on one component 
$\theta_i$ of $\btheta$, and we omit the index $i$:
$\theta := \theta_i$, $\mu := \mu_i$, $\sigma = \sigma_i$,
$\alpha := \alpha_i$, $r_{\sigma}(\mu) := r_{i,\sigma_i}(\mu_i)$.
Also, we denote by $\sigma_{0}^2$ the variance
of $\alpha_i$. We have:
\begin{align*}
r_{\sigma}(\mu)
= \mathrm{KL}(\beta_{\mu, \sigma} \| \alpha)
= \frac{1}{2}\ln\left(\frac{\sigma_{0}^2}{\sigma^2}\right) + \frac{\sigma^2 + \mu^2}{2\sigma_{0}^2}
- \frac{1}{2} .
\end{align*}
Thus we have:
\begin{align*}
\lambda r_{\sigma}(\mu) 
= \frac{1}{2} \ln\left(\frac{\frac{\sigma_{0}^2}{\lambda}}{\sigma^2}\right) 
+ \frac{\sigma^2 + \mu^2}{2 \frac{\sigma_{0}^2}{\lambda}} 
- \frac{1}{2} + K_{\lambda} ,
\end{align*}
where $K_{\lambda}$ does not depend on $(\mu, \sigma)$, so
its value does not affect optimization. Therefore:
\begin{align}
\lambda r_{\sigma}(\mu) = \mathrm{KL}(\beta_{\mu, \sigma} \| \alpha_{\lambda}),
\qquad \text{where} \qquad \alpha_{\lambda} \sim \mathcal{N}\left(0, \frac{\sigma_{0}^2}{\lambda}\right) . 
\label{eqn:alpha_lambda}
\end{align}

\paragraph{Step 2: assumption on the variance of the prior.}
Now that we have translated the penalty into a prior $\alpha_{\lambda}$, 
we are looking for a constraint to impose on $\alpha_{\lambda}$.
To do so, we make the following \emph{informal} postulate:
\begin{postulate} \label{postulate}
A distribution $\alpha$ over the parameters $\btheta$ of a neural network is a reasonable prior
over $\btheta$ if $\alpha$ can be used as an initialization distribution of $\btheta$.
\end{postulate}
\begin{proof2}
We consider a neural network (NN) before training.
In this situation, from a Bayesian point of view, 
no data point has been observed, so the Bayesian posterior
over the parameters is equal to the prior.

When we start training the NN with a batch of data,
we expect that the variational parameters $\mathbf{u}$ would move to the optimum.
Since we use the reparameterization trick \citep{kingma2014auto}
to train $\mathbf{u}$, we expect that the gradient of the loss according to
the parameters $\btheta$ of the NN is such that training is not stuck.

So, when we send the first batch of data to the NN for training,
we sample $\btheta$ from the prior $\alpha$,
and information should propagate and backpropagate properly
across the network. This condition corresponds to the ones
that are usually imposed on initialization distributions
\citep{glorot2010understanding,he2015delving,schoenholz2016deep}.

Therefore, to ensure that training is not stuck at the beginning,
it is reasonable to require the prior distribution to be usable
as an initialization distribution for $\btheta$.
\end{proof2}

This postulate has previously been used by \cite{ollivier2018online}.

More specifically, for the distribution of the weight $W$ of a neuron with $p$ inputs
and activation function $\varphi$:
\begin{itemize}
\item \cite{glorot2010understanding} recommend: 
$\mathrm{Var}(W) = \frac{1}{p}$ 
(under the hypothesis $\varphi = \mathrm{Id}$);
\item \cite{he2015delving} recommend: $\mathrm{Var}(W) = \frac{2}{p}$, when 
$\varphi = \mathrm{ReLU} = \max(\cdot, 0)$;
\item \cite{NEURIPS2019_9015} recommend: $\mathrm{Var}(W) = \left(\frac{5}{3}\right)^2 \frac{1}{p}$,
when $\varphi = \tanh$.
\end{itemize}
Now, we are ready to compute a reasonable $\lambda$.

\paragraph{Step 3: computation of $\lambda$.}
Regardless of the activation function $\varphi$, the initialization distribution
of a weight $W$ of a neuron with $p$ inputs has a variance:
$\sigma_*^2 = \frac{g^2}{p}$, where $g>0$ is a ``gain'' depending on $\varphi$.

According to Postulate \ref{postulate}, 
the variance of the prior $\alpha_{\lambda}$ should be equal to $\sigma_*^2$.
So, according to Eqn.~\eqref{eqn:alpha_lambda}, we should have:
\begin{align}
\frac{\sigma_{0}^2}{\lambda} = \frac{g^2}{p}. \label{eqn:heuristic}
\end{align}
Now, we just have to tune $\sigma_{0}^2$ and $\lambda$ so that Eqn.~\eqref{eqn:heuristic} holds.
Our choice is:
\begin{align}
\sigma_{0} = \frac{1}{\sqrt{p}} \qquad \text{and} 
\qquad \lambda = \frac{1}{g^2} . \label{eqn:heuristic1}
\end{align}

\subsection{Experimental setup}

\paragraph{Loss and priors.}
Above all, we average the loss \eqref{eqn:loss} over the
dataset $\mathcal{D}$ of size $n$:
\begin{align*}
\bar{\mathcal{L}}(\beta_{\mathbf{u}}) := \underbrace{- \frac{1}{n} \mathbb{E}_{\btheta \sim \beta_{\mathbf{u}}} \ln
p_{\btheta}(\mathcal{D})
}_{\text{NLL term}} \,+\, 
\underbrace{ \bar{\lambda} r(\mathbf{u})}_{\text{penalty term}},
\end{align*}
where $\bar{\lambda} = \frac{\lambda}{n}$ and 
``NLL'' stands for ``negative $\log$-likelihood''.
Our heuristic \eqref{eqn:heuristic1} becomes:
\begin{align}
\sigma_{0} = \frac{1}{\sqrt{p}} \qquad \text{and}
\qquad \bar{\lambda} = \frac{1}{n g^2} . \label{eqn:heuristic2}
\end{align}

We minimize $\bar{\mathcal{L}}$ according to the means and variances 
$(\mu_i, \sigma_i^2)$ of each parameter $\theta_i$.
The penalty associated with $(\mu_i, \sigma_i^2)$ is equal to 
$\frac{1}{n} \mathrm{KL}(\beta_{\mu_i, \sigma_i} \| \alpha_i)$, where $\alpha_i \sim \mathcal{N}(0, \frac{1}{p_i})$
and $p_i$ is the number of inputs of the neuron to which the weight $\theta_i$ belongs.

\paragraph{Datasets and architectures.}
We consider three setups:
\begin{enumerate}
\item MNIST + multilayer perceptron (MLP): layers of sizes $200, 100, 10$, $\varphi = \mathrm{ReLU}$;
\item CIFAR-10 + LeNet \citep{lecun1998gradient}, $\varphi = \tanh$;
\item CIFAR-10 + VGG13${}^*$%
\footnote{VGG13${}^*$ is VGG13 without batch-norm and with only one fully-connected layer.} 
\citep{simonyan2014very}, $\varphi = \mathrm{ReLU}$.
\end{enumerate}
In each setup, we train the variational parameters of the given architecture
with Adam \citep{kingma2015adam} and with various values for $\bar{\lambda}$,
centered in $\bar{\lambda} = \frac{1}{n}$.
We extracted a validation set from the usual training set,
of size 6000 for MNIST and of size 5000 for CIFAR-10. No data augmentation has been used.

\paragraph{Training details:}
\begin{enumerate}
\item MNIST + MLP: Adam with learning rate $\eta = 10^{-3}$, 200 epochs;
\item CIFAR-10 + LeNet: Adam with learning rate $\eta = 10^{-3}$, 400 epochs;
\item CIFAR-10 + VGG13${}^*$: Adam with learning rate $\eta = 10^{-4}$, 600 epochs.
\end{enumerate}

\subsection{Results}

\begin{figure}
\begin{subfigure}{.30\linewidth}
	\includegraphics[width=.95\linewidth]{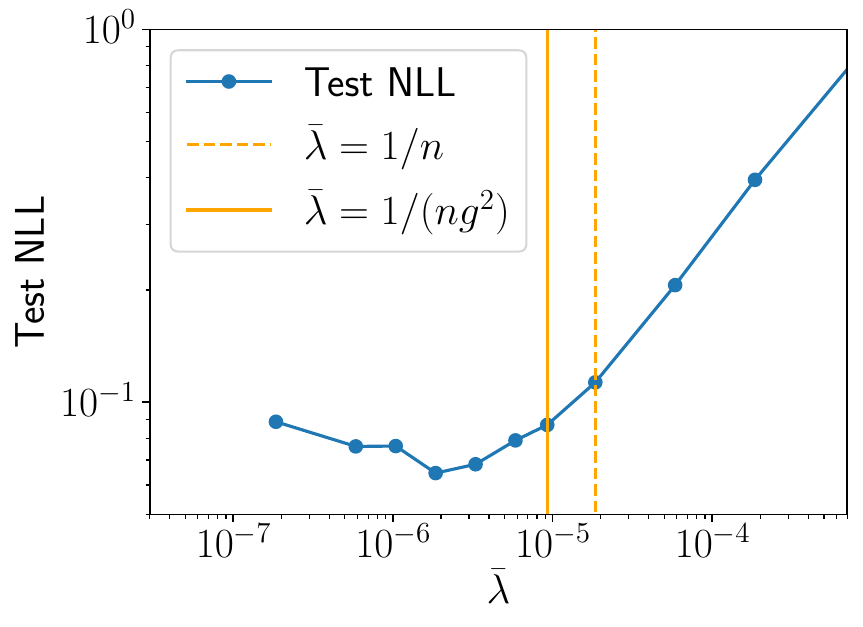}
	\subcaption{MNIST + MLP}
	\label{fig:main:a}
\end{subfigure}
\begin{subfigure}{.30\textwidth}
	\includegraphics[width=\linewidth]{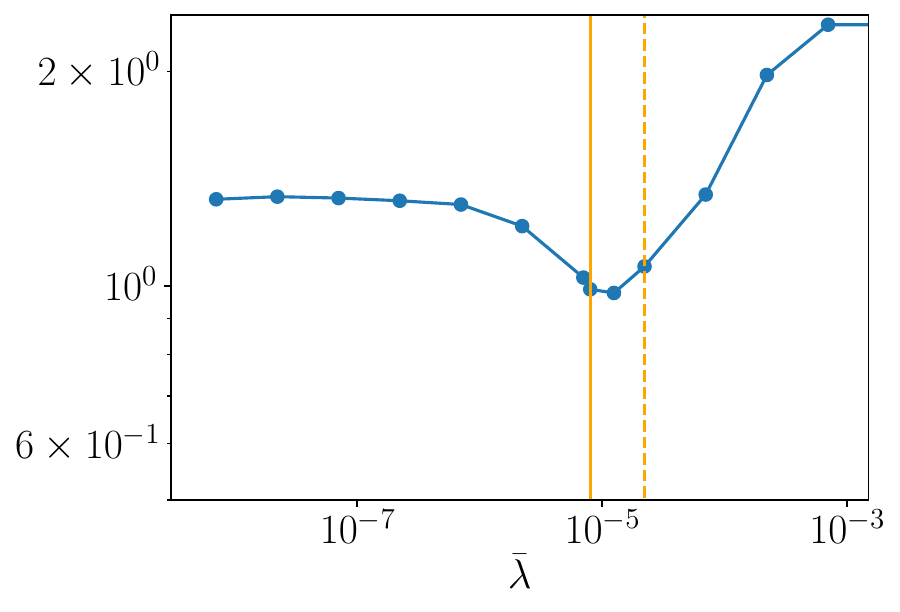}
	\subcaption{CIFAR-10 + LeNet}
	\label{fig:main:b}
\end{subfigure}
\begin{subfigure}{.30\textwidth}
	\includegraphics[width=\linewidth]{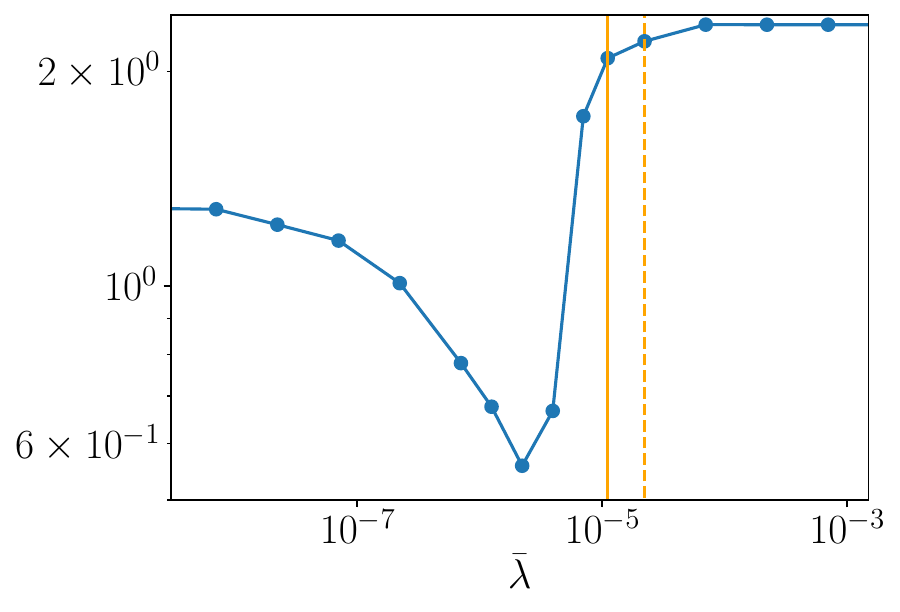}
	\subcaption{CIFAR-10 + VGG13${}^*$}
	\label{fig:main:c}
\end{subfigure}
\caption{Test NLL obtained at the epoch where the validation NLL is optimal, for 
	various penalty factors $\bar{\lambda}$. 
	The quality of our heuristics can be estimated by measuring the closeness
	of the minimum of the blue curve to the solid orange line.} \label{fig:main}
\end{figure}

We have plotted in Figure \ref{fig:main} the results for the 3 different setups.
Without surprise, we observe a transition between the overfitting regime
(small $\bar{\lambda}$) and the underfitting regime (large $\bar{\lambda}$):
when $\bar{\lambda}$ is small, the penalty term is too small to regularize
correctly the loss, and the model tends to overfit the training dataset;
when $\bar{\lambda}$ is large, the penalty term is too restrictive and prevents
the model from fitting correctly the data.

\paragraph{Correct order of magnitude.}
Between these regimes, there is some ``sweet spot'' for $\bar{\lambda}$, which is
typically the region where our heuristic should lie.
Let us denote our heuristic by $\hat{\lambda}$ and the optimal $\bar{\lambda}$
by $\lambda^*$.
In Figures \ref{fig:main:a} and \ref{fig:main:c}, our heuristic $\hat{\lambda}$ 
overestimates $\lambda^*$, while it slightly underestimates it
in Figure \ref{fig:main:b}.
Overall, our estimation does not deviate from the optimum by a factor $10$.
So, its order of magnitude is roughly correct.

\paragraph{Influence of the architecture.}
Interestingly, the error between our estimation and $\lambda^*$ is much greater 
in Figure \ref{fig:main:c} than in Figure \ref{fig:main:b}, while the 
$\hat{\lambda}$ is approximately the same in both cases.
Two factors explain this similarity between the $\hat{\lambda}$:
the size of the training dataset is the same in both cases; 
the activation functions used in both cases have a similar squared gain $g^2$.
However, $\hat{\lambda}$ lies in the sweet spot in Figure \ref{fig:main:b},
and is clearly outside of it in Figure \ref{fig:main:c}.

Consequently, a good heuristic for $\bar{\lambda}$ cannot
rely only on the size of the training dataset and the choice of activation function.
One can assume that the quality of our proposal $\hat{\lambda}$ depends
also on the architecture of the NN, including its number of
parameters (LeNet has only $60~000$ parameters, while VGG13${}^*$ has roughly $10~000~000$
parameters).
This empirical finding corroborates \citep{izmailov2021bayesian},
which states that the prior design should take into account 
the architecture of the NN, via the function mapping the vector of parameters
of a NN to the function represented by the NN.

\paragraph{Link with the cold posterior effect.}
According to \cite{wenzel2020good}, the ``cold posterior effect'' is
an empirical finding according to which, when training neural networks,
a posterior tempered with a well-chosen temperature $T < 1$ leads to a better
test loss than the Bayesian (non-tempered, $T = 1$) posterior.
This finding is related to the choice of the penalty factor $\bar{\lambda}$:
if $T < 1$, then $\bar{\lambda} < 1$.

However, as shown in Eqn.~\eqref{eqn:heuristic}, the quantity of interest is not
$\bar{\lambda}$ alone, but the ratio $\nicefrac{\sigma_{0}^2}{\lambda}$.
So, focusing only on the optimal value of $\bar{\lambda}$ is meaningless if
the variance $\sigma_{0}^2$ is chosen carelessly.

We note that many works about Bayesian inference applied to NNs,
such as \citep{zhang2018noisy,bae2018eigenvalue,ashukha2019pitfalls},
but also \citep{izmailov2021bayesian}, which obtains posteriors of much better quality
than with VI by using HMC and \citep{wenzel2020good},
assume that the posterior is isotropic. 
This assumption is out of step with our heuristic (see Eqn.~\eqref{eqn:heuristic} and
\eqref{eqn:heuristic2}): 
if we want the prior distribution $\alpha$ to be usable to initialize the NN, then
\emph{$\alpha$ cannot be isotropic}.%
\footnote{\cite{fortuin2022bayesian} have considered non-isotropic multivariate priors,
but not non-isotropic priors with diagonal covariance, which would be close to what 
we are doing.}
The variance of the prior over a weight $\theta_i$ depends on
some architecture hyperparameters (number of neurons per layer).
This choice of diagonal layer-dependent prior covariance has previously been proposed
by \citep{ollivier2018online}.

Nonetheless, Figures \ref{fig:main:a} and \ref{fig:main:c} exhibit
optimal values $\lambda^*$ that are smaller than $\hat{\lambda}$,
which seems to confirm the ``cold posterior effect''.
But, again, one could argue that our choice of prior is not entirely satisfying
and propose a better heuristic for $\alpha$, which would 
consequently change the heuristic for $\lambda$, challenging again the
``cold posterior effect''.

\section{Discussion} \label{sec:discussion}

\paragraph{Conditions of Theorem \ref{thm:main}.}
Overall, the technical conditions of Theorem \ref{thm:main} do not play the same role.
Condition \ref{cond:a:schwartz} ensures that the $\mathrm{KL}$-divergence in Eqn.~\eqref{eqn:thm:1}
remains computable for a very wide range of $\log$-priors $A$.
Similarly, Condition \ref{cond:c:quotient} ensures that $A$ can be computed 
within the framework of the standard theory of distributions. 
Specifically, the Fourier transform of $\nicefrac{\mathcal{F} r_{\bnu}}{\mathcal{F} \check{\beta}_{0, \bnu}}$
can be computed.
Finally, Condition \ref{cond:b:nonzero} ensures the uniqueness of the solution of Eqn.~\eqref{eqn:thm:1}
in $\mathcal{S}'(\mathbb{R}^N)$.

\paragraph{Trade-off between the Condition \ref{cond:a:schwartz} and the search space $\mathcal{T}$.}
We have chosen a very strong Condition \ref{cond:a:schwartz}, 
that is, each $\beta_{\bmu, \bnu}$ should belong to $\mathcal{S}(\mathbb{R}^N)$,
in order to allow for $\log$-priors $A$ to belong to the very large space $\mathcal{S}'(\mathbb{R}^N)$.
If we want to deal with variational families with less regular densities%
\footnote{For instance, derivable only a finite number of times and 
with a polynomial decrease at infinity.},
we would have to shrink the search space of $A$.

%

\paragraph{Allowing improper priors.}
In Section \ref{sec:appl:cosine_penalty}, the solution $\alpha$ to Eqn.~\eqref{eqn:cor:1}
is an improper prior.
If one wants a proper prior, one can add an $\mathcal{L}^2$ penalty
to the initial cosine penalty, which would impose Gaussian tails.
But, it is also possible to deal with improper priors in general,
if we are certain that the resulting variational posterior is 
a probability distribution.
To allow improper priors, one just needs to relax Corollary \ref{cor:main}
by removing Condition \ref{cond:d':proper}.


\paragraph{Beyond translation-invariant variational families.}
To prove Theorem \ref{thm:main}, we have used the fact that the 
variational family $(\beta_{\bmu, \bnu})_{\bmu, \bnu}$ 
is translation-invariant: 
$\beta_{\bmu, \bnu}(\btheta) = \beta_{0, \bnu}(\btheta - \bmu)$ for
all $\bmu, \bnu, \btheta$.
This condition allowed us to formulate a part of the $\mathrm{KL}$
as a convolution between two functions, which can be disentangled
by using the Fourier transform:
$\mathcal{F}_{\bmu} \int_{-\infty}^{\infty} A(\btheta) \beta_{\bmu, \bnu}(\btheta) \, \mathrm{d}\btheta 
= (\mathcal{F} A) \cdot (\mathcal{F} \check{\beta}_{0, \bnu})$.

But there are other ways to disentangle functions inside an integral,
if we assume different properties for the variational family.
For instance, one can deal with a \emph{scale-invariant} variational family:
$\beta_{\mu, \bnu}(\theta) = \frac{1}{\mu}\beta_{1, \bnu}(\frac{\theta}{\mu})$ for
all $\mu, \bnu, \theta$, and with support included in $\mathbb{R}^+$.
In this case, we can express the $\mathrm{KL}$ with a \emph{multiplicative convolution} $\star$,
which can be disentangled by using the \emph{Mellin transform} $\mathcal{M}$:
first, transform the cross-entropy term into a multiplicative convolution:
$\int_{0}^{\infty} A(\theta) \beta_{\mu, \bnu}(\theta) \, \mathrm{d}\theta 
= \frac{1}{\mu}\int_{0}^{\infty} \theta A(\theta)
\tilde{\beta}_{1, \bnu}(\mu/\theta) \, \frac{\mathrm{d}\theta}{\theta}
= \frac{1}{\mu} [(\mathrm{Id} \cdot A) \star \tilde{\beta}_{1, \bnu}](\mu)$,
where $\mathrm{Id}$ is the identity function
and $\tilde{\beta}(\theta) = \beta(1/\theta)$;
second, apply the Mellin transform $\mathcal{M}$:
$\mathcal{M}_{\mu} [\mu \int_{0}^{\infty} A(\theta) \beta_{\mu, \bnu}(\theta) \, \mathrm{d}\theta] 
= (\mathcal{M} [\mathrm{Id} \cdot A]) \cdot (\mathcal{M} \tilde{\beta}_{1, \bnu})$;
third, use basic properties of $\mathcal{M}$: 
$(\mathcal{M} \int_{0}^{\infty} A(\theta) 
\beta_{\cdot, \bnu}(\theta) \, \mathrm{d}\theta )(t + 1)
= (\mathcal{M} A)(t + 1) \cdot (\mathcal{M} \beta_{1, \bnu})(-t)$.
Finally, it is possible to isolate $\mathcal{M}A$ and recover
$A$ by using $\mathcal{M}^{-1}$.
Then, one can rewrite Theorem \ref{thm:main} and Corollary \ref{cor:main}
with the appropriate formulas and hypotheses.

\paragraph{Heuristics for the prior and the penalty factor in the case of
neural networks.}
In Section \ref{sec:application},
we showed how to combine our main theoretical results with usual initialization heuristics
in neural networks to discover a heuristic for the prior and the penalty factor.
We showed the results of such heuristics in a series of experiments, 
and we proposed a discussion about the ``cold posterior effect'' \citep{wenzel2020good}
in variational inference. Notably, we suggest a specific choice of the prior
distribution, which should be non-isotropic, and with a variance over the weights 
of each layer depending on the input dimension of the considered layer.

\section{Conclusion}

We have provided a theorem that bridges the gap between empirical penalties and Bayesian priors when learning 
the distribution of the parameters of a model through Variational Inference.
This way, various regularization techniques can be studied in a single Bayesian framework, and be 
seen as probability distributions. 
This unification of points of view is valuable on the theoretical side, since it can be used to apply 
well-know theorems in variational inference, such as results about convergence rates. 
Also, we have shown experimentally material that such theoretical results can be
used in practice to help tuning the hyperparameters, especially when the model exhibits some structure, such
as neural networks.

\bibliography{PenaltyFactor_2024_02_arXiv}
\bibliographystyle{apalike}

\end{document}